\documentclass[preprints,article,accept,moreauthors,pdftex]{Definitions/mdpi} 
\firstpage{1} 
\pubvolume{xx}
\issuenum{1}
\articlenumber{5}
\pubyear{2019}
\copyrightyear{2019}
\history{Received: date; Accepted: date; Published: date}

\usepackage{color}
\usepackage{todonotes}
\usepackage[flushleft]{threeparttable}

\usepackage{fancyhdr,graphicx,amsmath,amssymb}
\usepackage[ruled,vlined]{algorithm2e}

\Title{Subsentence extraction from text using coverage-based deep learning language models}

\def\fps@figure{htp}
\def\fps@table{htp}

\newcommand{\bi}{\begin{itemize}}
\newcommand{\ei}{\end{itemize}}

\newcommand{\bfig}{\begin{figure}}
\newcommand{\efig}{\end{figure}}

\newcommand{\benum}{\begin{enumerate}}
\newcommand{\eenum}{\end{enumerate}}

\newcommand{\be}{\begin{equation}}
\newcommand{\ee}{\end{equation}}

\newcommand{\ba}{\begin{eqnarray}}
\newcommand{\ea}{\end{eqnarray}}

\newcommand{\unit}[1]{\mbox{$\rm \,#1$}}

\Author{JongYoon Lim $^{1,\dagger,\ddagger}$\orcidA{}, Inkyu Sa $^{2,\ddagger}$*\orcidA{0000-0001-5429-0515}, Ho Seok Ahn$^{1}$\orcidA{}, Norina Gasteiger$^{1}$\orcidA{}, Sanghyub John Lee$^{1}$\orcidA{}, and Bruce MacDonald$^{1}$\orcidA{}}

\AuthorNames{JongYoon Lim, Inkyu Sa, Ho Seok Ahn, Norina Gasteiger, Sanghyub John Lee, and Bruce MacDonald}

\address{%
$^{1}$ \quad CARES, Department of Electrical, Computer and Software Engineering, University of Auckland; {jy.lim, hs.ahn, n.gasteiger, sanghyub.lee, b.macdonald}@auckland.ac.nz\\
$^{2}$ \quad CSIRO Data61, Robotics and Autonomous Systems Group, Dynamic platforms
; inkyu.sa@csiro.au}

\corres{Correspondence: inkyu.sa@csiro.au; Tel.: +61 7 3327 4317 (S. I.)}

\firstnote{Current address: Affiliation 3} 
\secondnote{These authors contributed equally to this work.}

\abstract{
Sentiment prediction remains a challenging and unresolved task in various research fields, including psychology, neuroscience and computer science. This stems from its high-degree of subjectivity and limited input sources that can effectively capture the actual sentiment. This can be even more challenging with only text-based input. Meanwhile, the rise of deep learning and an unprecedented large volume of data have paved the way for artificial intelligence to perform impressively accurate predictions or even human-level reasoning. Drawing inspiration from this, we propose a coverage-based sentiment and subsentence extraction system that estimates a span of input text and recursively feeds this information back to the networks. The predicted subsentence consists of auxiliary information expressing a sentiment. This is an important building block for enabling vivid and epic sentiment delivery (within the scope of this paper) and for other natural language processing tasks such as text summarisation and Q\&A. Our approach outperforms the state-of-the-art approaches by a large margin in subsentence prediction (i.e., Average Jaccard scores from 0.72 to 0.89). For the evaluation, we designed rigorous experiments consisting of 24 ablation studies. Finally, our learned lessons are returned to the community by sharing software packages and a public dataset that can reproduce the results presented in this paper.}


\begin{document}



\section{Introduction}

Understanding human emotion or sentiment is one of the most complex and active research areas in physiology, neuroscience, neurobiology, and computer science. Liu and Zhang \cite{liu2012survey} defined sentiment analysis as the computational study of people’s opinions, appraisals, attitudes, and emotions toward entities, individuals, issues, events, topics and their attributes. The importance of sentiment analysis is gaining momentum in both individual and commercial sectors. Customers' patterns and emotional states can be derived from many sources of information (e.g., reviews or product ratings), and precise analysis of them often leads to direct business growth. Sentiment retrieval from personal notes or extracting emotions in multimedia dialogues generates a better understanding of human behaviours (e.g., crime, or abusive chat). Sentiment analysis and emotion understanding have been applied to the field of Human-Robot Interaction (HRI) in order to develop robotics that can form longer-term relationships and rapport with human users by responding with empathy. This is crucial for maintaining interest in engagement when the novelty wears off. Additionally, robots that can identify human sentiment are more personalized, as they can adapt their behaviour or speech according to the sentiment. The importance of personalization in HRI is widely reported and impacts cooperation \cite{Lee2012}, overall acceptability \cite{Clabaugh2019, Di2018} and interaction and engagement \cite{Lee2012, Henkemans2013}. Indeed, an early review by Fong \cite{Fong2003} on characteristics of successful socially interactive robots suggested that emotions, dialogue and personality are crucial to facilitating believable HRI.

Sentiment can be efficiently detected by exploiting multimodal sensory information such as voice, gestures, images or heartbeats, as evident in well-established literature and studies in physiology and neurosciences \cite{Acheampong2020-al}. However, reasoning sentiment from text information such as tweets, blog posts and product reviews is a more difficult problem. This is because text often contains limited information to describe emotions sufficiently. It also differs across gender, age and cultural backgrounds \cite{ahn2012uses}. Furthermore, sarcasm, emojis and rapidly emerging new words restrict sentiment analysis from text data. Conventional rule-based approaches may fail in coping with these dynamic changes.

\begin{figure}
\centering
\includegraphics[width=\textwidth]{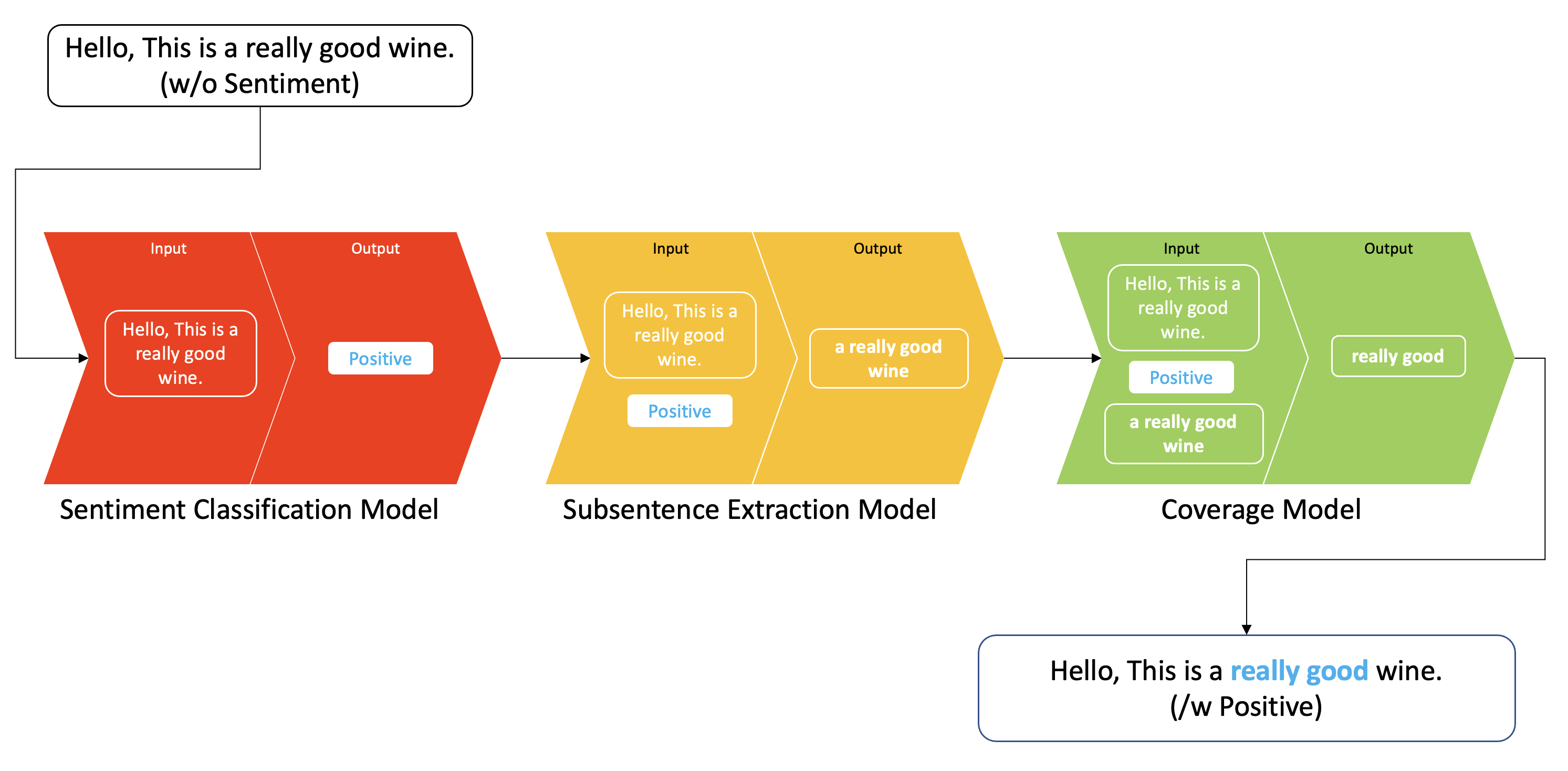}
\caption{Processing pipeline with an example input sentence. There are three cascaded models to predict both sentiment and subsentence. Each model consists of a bidirectional transformer model with varying output layers.}
\label{fig:pipeline}
\end{figure}

To address these challenges, we present a study that adopts the latest technologies developed in artificial intelligence (AI) and natural language processing (NLP) for sentiment and subsentence extraction, as shown in Figure~\ref{fig:pipeline}. The recent rise of AI with deep neural networks and massively big data have demonstrated super human-level performance in many tasks including image classification and object detection, cyber-security, entertainment (e.g., playing GO or DOTA), and NLP (e.g., Q\&A \cite{brown2020language}). Sentiment detection from text is one of the subsets of NLP. It is convincing that these language models can adequately capture the underlying sentiment from text data (more details are presented in section \ref{sec:related work}).


This paper presents a sentiment classification and subsentence extraction model from text input using deep neural networks and shows an improvement of the extraction accuracy with a coverage-based span prediction model. Therefore the contributions of this paper are;
\begin{itemize}
    \item Proposing a novel coverage-based subsentence extraction system that takes into considerations the length of subsentence, and an end-to-end pipeline which can be directly leveraged by Human-Robot Interaction
    \item Performing intensive qualitative and quantitative evaluations of sentiment classification and subsentence extraction
    \item Returning lessons and learnt to the community as a form of open dataset and source code\footnote{\url{https://github.com/UoA-CARES/BuilT-NLP}}.
\end{itemize}

The rest of the paper is structured as follows;
Section \ref{sec:related work} presents the State-Of-The-Art (SOTA) NLP studies in sentiment analysis and its relevant usage within HRI. Section \ref{sec:method} addresses the detailed approach we proposed, such as coverage-based subsentence extraction and Exploratory Data Analysis (EDA) of the dataset. Section \ref{sec:dataset} delineates the dataset used in this paper and its preparation for model training that is addressed in section \ref{sec:modelTraining}. This is followed by experiments results in sentiment classification and subsentence extraction in section \ref{sec:results}. We discuss the advantages and limitations of the proposed approach in section \ref{sec:discuss} to highlight opportunities for future work in this field and conclude the paper by presenting a summary in section \ref{sec:conclusions}.

\section{Related work}
\label{sec:related work}

Sentiment detection is an interdisciplinary task that requires tight connections between multiple disciplines. In this section, we introduce related studies in NLP, sentiment analysis and emotion detection, and HRI.

\subsection{Natural Language Processing for sentiment analysis}
There has been a large volume of public or commercial interests in NLP for many decades. The primary objective is to aid machines in understanding human languages by utilising linguistics and computational techniques \cite{Pennington2014-cy}. Especially, its importance and impact are gaining momentum with the rise of big data and deep learning techniques in interpreting human-level contextual information or generating vivid artificial articles (Generative Pre-trained Transformer 3, GPT-3) \cite{brown2020language}. Nowadays, the prominent applications of NLP are language translation \cite{johnson2017google}, text summarisation \cite{cui2016attention}, Q\&A task \cite{soares2020literature,rajpurkar2018know,rajpurkar2016squad, Zhou_undated-lb,wang2018glue, lai2017race}, information retrieval \cite{Crnic2011-yz}, and sentiment analysis \cite{cambria2017affective, Kanakaraj2015-ig, guo2019novel}. 

Most of these data-driven approaches require a high fidelity and quality training dataset. There are several publicly available datasets as reported by \cite{Acheampong2020-al}. These datasets are valuable and cover broad spectrums such as cross-cultural studies, news, tales, narratives, dialogues, and utterances. However, all these datasets only provide an emotion label for each sentence rather than fine-grain subsentence labels that we leverage in this paper. For instance, these datasets only provide positive sentiment for a sentence "Hello this is a really good wine", whereas we aim to predict both polarity of the sentence and the parts of a sentence which lead to the positive sentiment; "really good". In this context, to the author's best knowledge, the dataset we used is unique and one of the pioneering studies in processing fine-grain sentiment detection.

\subsection{Emotion detection in sentiment analysis}
Emotion detection is a subset of sentiment analysis that seeks not only polarity (e.g., positive, negative, or neutral) from input sentence or speech but tries to derive more detailed emotions (e.g., happy, sad, anxious, nervous instead positive or negative). Regarding this topic, there is an interesting survey report \cite{Acheampong2020-al}. This comprehensive survey focuses on conventional rule-based and current deep learning-based approaches. There are two key points from this article; it highlights the use of the bidirectional transformer model \cite{devlin2018bert} that boosts overall classification performance by a large margin and it lacks adequate application of emotion detection. These points are well aligned with our approach and goal; we used a variant Bidirectional Encoder Representations from Transformers (BERT) language model \cite{liu2019roberta} as a backbone model and proposed a practical application in HRI. This may reflect that the approach and the research goal of this paper are well-defined. We believe that the proposed method can be exploited in many relevant domains such as daily human-machine interaction (e.g., Alexa or Google Assistant), mental care services (e.g., depression treatment), or in aged care.

Sentiment analysis is a sophisticated interdisciplinary field that connects linguistic and text mining with AI, predominantly for NLP \cite{Sza2020}. Dominant sentiment analysis strategies may include emotion detection, subjectivity detection or polarity classification (e.g., classifying positive, negative and neutral sentiments \cite{de2012sentisense}). These techniques require an understanding of complicated human psychology. For instance, they may include instant and short-lasting emotions \cite{ekman1984expression}, longer-lasting and unintentional moods \cite{amado1993concept}, and feelings or responses to emotions \cite{Sza2020}.

\subsection{Human-Robot Interaction}
The importance of personalisation in HRI is widely reported, and some experiments have been conducted to compare robotic systems with and without sentiment abilities, including expressing and analysing sentiment \cite{Breazeal2003, striepe2017there, Sza2020, shen2015sentiment, bae2012towards}. These studies used the NAO, Kismet and Reeti robots. An early and well-known study by Breazeal \cite{Breazeal2003} recruited five participants to interact with Kismet in different languages. The individuals understood the robot's emotions successfully. Additionally, humans and the robot mirrored one another, demonstrating a phenomenon called \textit{affective mirroring}. This psychological phenomenon is often observed in human-to-human interaction, whereby an emotion is subconsciously mirrored. This was evidenced in the experiment, whereby participants apologised and looked to the researcher with anguish, displaying guilt or stating that they felt terrible when they made Kismet feel sad. 

In three experimental studies, the robots were evaluated when completing the task of reading fables/stories. The purpose of these experiments was to determine the effect of emotion in HRI. In one example, Striepe \cite{striepe2017there} recruited 63 German participants aged 18 to 30 years and assigned them to one of three groups: audiobook story, emotional social robot storyteller or neutral robot storyteller. They concluded that the emotional robot storyteller was just as effective at ‘transporting’ listeners into the story as the audiobook. The neutral robot performed the worst. Similar findings were found by Szabóová \cite{Sza2020} who reported that their robot with emotion expression, for example, voice pitch, was better rated overall. Participants even thought that the robot appeared capable of understanding, compared to the neutral robot. These findings also transferred to other activities, such as gaming. Shen \cite{shen2015sentiment} developed a robot that mimicked user’s facial expressions and one that demonstrated sentiment apprehension, for example, a robot's ability to reason about the user's attitudes such as judgment/liking. They found that people wanted to play games on the robot with sentiment apprehension more than the other, as well as rated it to be more engaging. Evidently, sentiment plays an important role in HRI, including in simulating human phenomena and perceptions, for example, understanding and affective mirroring and being superior to their neutral counterparts.

Some observational (non-experimental) research has also explored the effectiveness of emotion expression by robots. For example, in Rodriguez \cite{rodriguez2017adaptive} 26 nine-year-old children interacted with a robot to determine whether they could understand the robot’s emotional expression. The authors concluded that the children could understand when the robot was expressing sad or happy emotions, by its facial expressions and gestures (e.g., fast and slow movements or changes of lights to its eyes). Additionally, in a small pilot study with four participants, an emotionally expressive storytelling robot attempted to persuade listeners to make a decision, using facial expressions and head movements \cite{paradeda2018would}. The authors conclude that an amended version of this robot may potentially increase the motivation, interest and engagement of users in the task.


From the literature review, we have shown that deep learning transformer language models' utilisation is one of the main streams in text-based sentiment extraction. Besides, there exists a massive demand for such systems in various applications mentioned above (e.g., HRI). However, there are still remaining challenges, such as the lack of a unified sentiment extraction pipeline that can be directly applied to existing systems and the need for a high-performance sentiment extraction model. 

We propose coverage-based subsentence extraction and an easy-to-use processing pipeline in the following sections to address these challenges.

\section{Methodologies}
\label{sec:method}

In this section, we present the methodologies used later in this paper, including Bidirectional Encoder Representations from Transformers (BERT) and A Robustly Optimized BERT Pretraining Approach (RoBERTa) NLP models followed by sentiment classification and coverage-based subsentence extraction.


\subsection{BERT and RoBERTa transformer language models}
\label{sec:bert-roberta}
Our proposed model is built on RoBERTa \cite{liu2019roberta}, which is one of the variants of BERT \cite{devlin2018bert} that improves overall performance (e.g., F1 scores) in language understanding, Q\&A tasks by taking into account of dynamic masking, full-sentences without Next Sentence Prediction (NSP) loss, and a larger byte-level Byte-Pair Encoding (BPE) with additional hyper-parameter tuning (e.g., batch size). With the superior performance of RoBERTa over BERT, another selection criteria is that it makes use of dynamic masking which can efficiently handle high-variance daily-life text such as tweets, text messages, or daily dialogues. Technical details and deeper explanations can be found in the original papers \cite{devlin2018bert, liu2019roberta}. We therefore only provide a summary and high-level overview for contextual purposes.

BERT is considered one of the most successful language models for NLP proposed by the Google AI language team in 2018. As the name implies, the fundamental idea is taking account of both sequential directions (i.e., left-to-right and right-to-left) in order to capture more meaningful contextual information. Inside of these models, an attention-driven language learning model, Transformer \cite{Vaswani2017-vv} (only the encoder part), is exploited in our method to extract context with Next Sentence Prediction (NSP) and Masked Language Model (MLM). Varying pre-trained BERT models exist depending on the number of encoders, such as BERT-base (12 encoders, 110\unit{M} parameters) or BERT-large (24 encoders, 330\unit{M} parameters), with the larger model often producing better results but taking more training time.

\begin{figure}
\centering
\includegraphics[width=\textwidth]{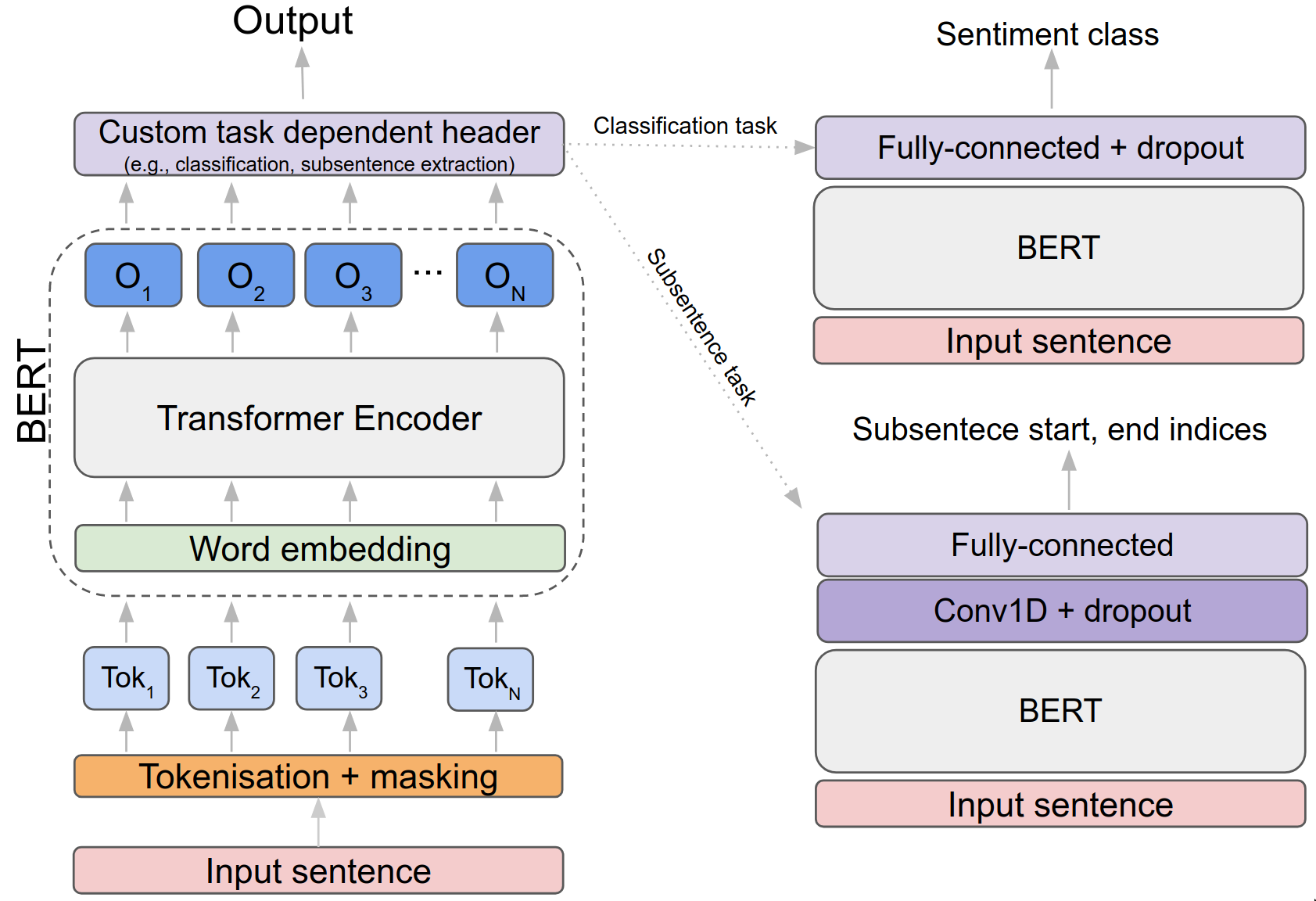}
\caption{An overview of BERT (left) and task-driven fine-tuning models (right). Input sentence is split into multiple tokens ($Tok_N$) and fed to a BERT model which outputs embedded output feature vectors, $O_N$, for each token. By attaching different head layers on top, it transforms BERT into a task-oriented model.}
\label{fig:bert}
\end{figure}

NSP plays a role in distinguishing whether two input sentences are highly correlated or not, by formulating a binary classification problem (e.g., if two adjacent sentences are sampled from the same dataset, this pair is treated as positive, otherwise negative samples). MLM is performed by randomly masking words from the input sentence with [MASK] (e.g., ``Today is a great day'' can be ``Today is a [MASK] day''), and the objective of MLM is the correct prediction of these masked words. Figure \ref{fig:bert} illustrates the overall pipeline and BERT's pre-training and fine-tuning procedures. For more technical detail, please refer to the original paper \cite{devlin2018bert}.

\begin{figure}
\centering
\includegraphics[width=\textwidth]{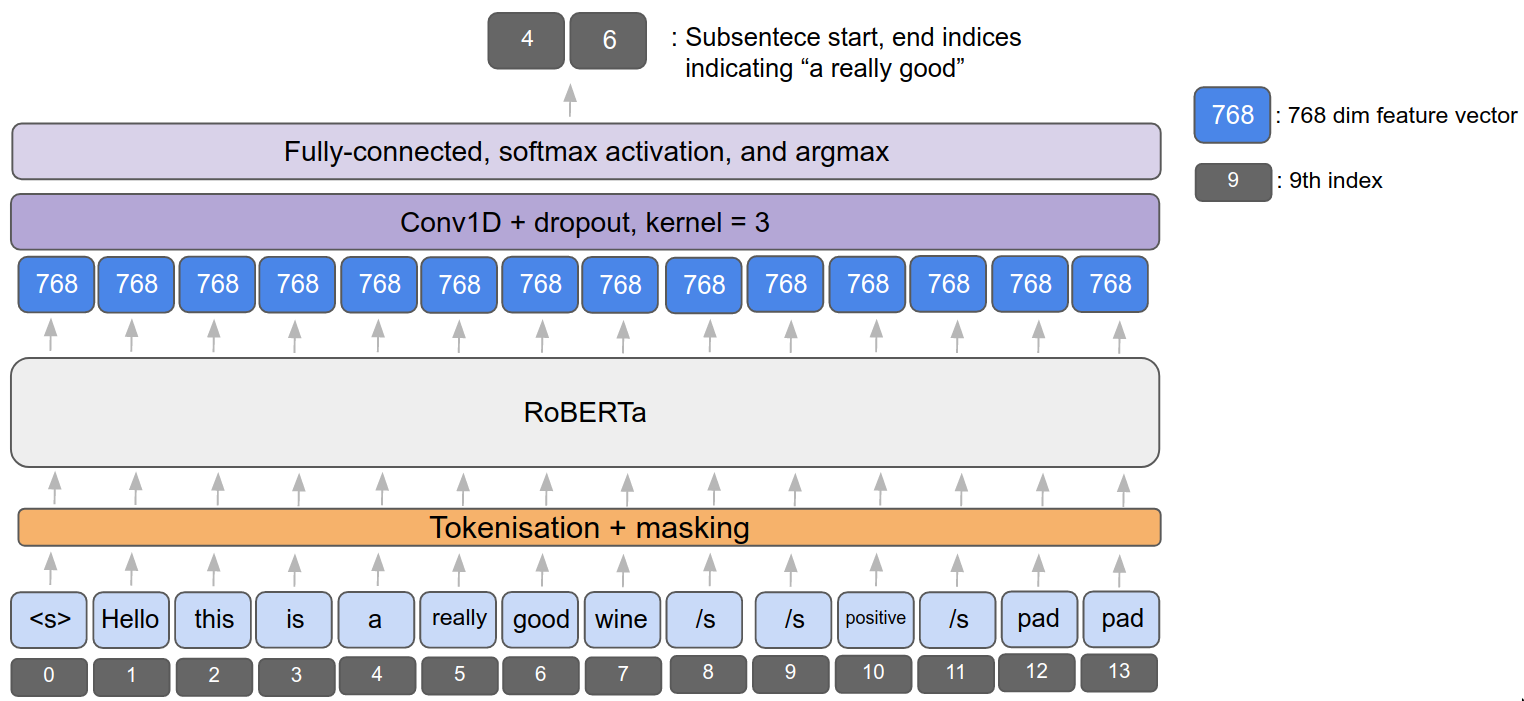}
\caption{This diagram exemplifies how input sentence is processed using RoBERTa model. Gray indicates a token index and blue is embedded output feature vector.}
\label{fig:roberta}
\end{figure}

RoBERTa proposed a couple of network design choices and BERT training strategies such that adopting dynamic masking and removing NSP, which may degrade the performance in specific downstream tasks. In addition to these, intensive hyper-parameter tuning and ablation studies were conducted with various tasks and public datasets. Dynamic masking is more suitable for our task dealing with high-variance datasets such as tweets or daily conversations. Drawing inspiration from this, we decided to build our baseline model upon RoBERTa.
Figure \ref{fig:roberta} illustrates sentiment inference task using a pre-trained $\text{BERT}_{\mbox{\tiny{base}}}$ model. A positive contextual sentence (i.e., tokenised, covered this in the next section) is fed into the model, creating 768 feature vectors for each token (by feature embedding). It is then processed throughout the custom headers such as 1D Convolution followed by Softmax activation. The final output then indicates that the subsentence from 4th to 6th token (i.e., \texttt{"a really good"}) is predicted from the network. 

\subsection{Sentiment Classification}
\label{sec:sentimentClassification}
Given the transformers model mentioned earlier, we are interested in extracting subsentences from an input sentence and estimating the sentiment of the sentence. It is mainly because this information can significantly boost the subsentence extraction prediction by narrowing down the potential search space. It is also useful for other applications such as tone estimation or emotion detection. This problem is well-formulated as supervised-learning, meaning that we provide a known label (e.g., positive, negative, or neutral) for a sentence. A neural network is asked to predict the corresponding label.

Although we only consider three classes in this paper, there are no limitations to the number of classes as long as the class label is provided. We design a simple neural network with two additional layers (i.e., dropout and Fully-connected layer) on top of the RoBERTa model). The overall sentiment classification network architecture is shown in Figure \ref{fig:bert} and the relevant results are presented in the section \ref{sec:classificaiton-results}.


\subsection{Coverage model for subsentence extraction}
\label{sec:coverage}
Transformer models have been widely used for NLP, image-based classification or object detection tasks \cite{dosovitskiy2020image}. These attention-based transformer networks can be improved by providing additional metadata. Sentiment information and attention mask, for example, are useful sources for subsentence extraction. Furthermore, extracting efficient features such as a span of subsentence or character-level encoding instead of word-level encoding can also improve the model's performance.

Within this context and drawing inspiration from \cite{wallace2019nlp}, we propose a recursive approach that estimates the length of subsentence in an input sentence. We refer the length of subsentence as coverage, $c$, is a scalar and computed as $c = \frac{M}{N}\times\kappa$ where $M$ and $N$ are the length of subsentence and input sentence respectively. $\kappa$ is a scale parameter governing the width of $c$. We empirically initialised this as 15. It is worth mentioning that the overall performance is more or less independent of this, if and only, if $c$ is larger than a threshold. This is because $c$ will decrease regardless of an initial value. However, we found that if $c$ was too small, then the model struggled to find the correct subsentence because our model could not expand the coverage range. Regarding this issue, we discuss more detail in the future work and limitations section~\ref{sec:discuss}.

In summary, our coverage model takes the following steps: 1) predicting the length of subsentence (i.e., coverage) by utilising a transformer network that outputs start and end indices (e.g., Q\&A or text summarisation networks). 2) Computing coverage and feeding back into a coverage model with an input sentence and previously predicted indices. For more detail and better understanding, algorithm~\ref{algo:covg} delineates model pseudo code.


\begin{algorithm}[H]
\SetAlgoLined
\SetKwInput{KwInput}{Input}
\SetKwInput{KwOutput}{Output}
\DontPrintSemicolon
\KwInput{sentence, sentiment}
\KwOutput{start\_idx, end\_idx}

\SetKwFunction{FGlobal}{}
\SetKwFunction{FgetPrediction}{getPrediction}
\SetKwFunction{FgetCoverage}{getCoverage}

  \SetKwProg{Fn}{Global variable}{:}{}
  \Fn{\FGlobal{}}{
        $\epsilon$ = 0.1\;
        text\_len = len(sentence)\;
  }
  \SetKwProg{Fn}{Def}{:}{}
  \Fn{\FgetPrediction{sentence, sentiment}}{
        s\_idx, e\_idx = classification\_model\_inference(sentence, sentiment)\;
        \KwRet s\_idx, e\_idx\;
  }
  
  \SetKwProg{Fn}{Def}{:}{}
  \Fn{\FgetCoverage{sentence, sentiment}}{
        pred\_s\_idx, pred\_e\_idx = getPrediction(sentence, sentiment)\;
        covg\_s\_idx, covg\_e\_idx = coverage\_model\_inference(pred\_s\_idx, pred\_e\_idx, sentiment)\;
        pred\_len = pred\_e\_idx - pred\_s\_idx\;
        \eIf{pred\_len / text\_len > $\epsilon$}{
          final\_s\_idx = covg\_s\_idx\;
          final\_e\_idx = covg\_e\_idx
          }{
          final\_s\_idx = pred\_s\_idx\;
          final\_e\_idx = pred\_e\_idx
        }
        \KwRet final\_s\_idx, final\_e\_idx\;
  }
 \caption{Coverage-based subsentence extraction algorithm}
 \label{algo:covg}
\end{algorithm}

\begin{figure}
\centering
\includegraphics[width=\textwidth]{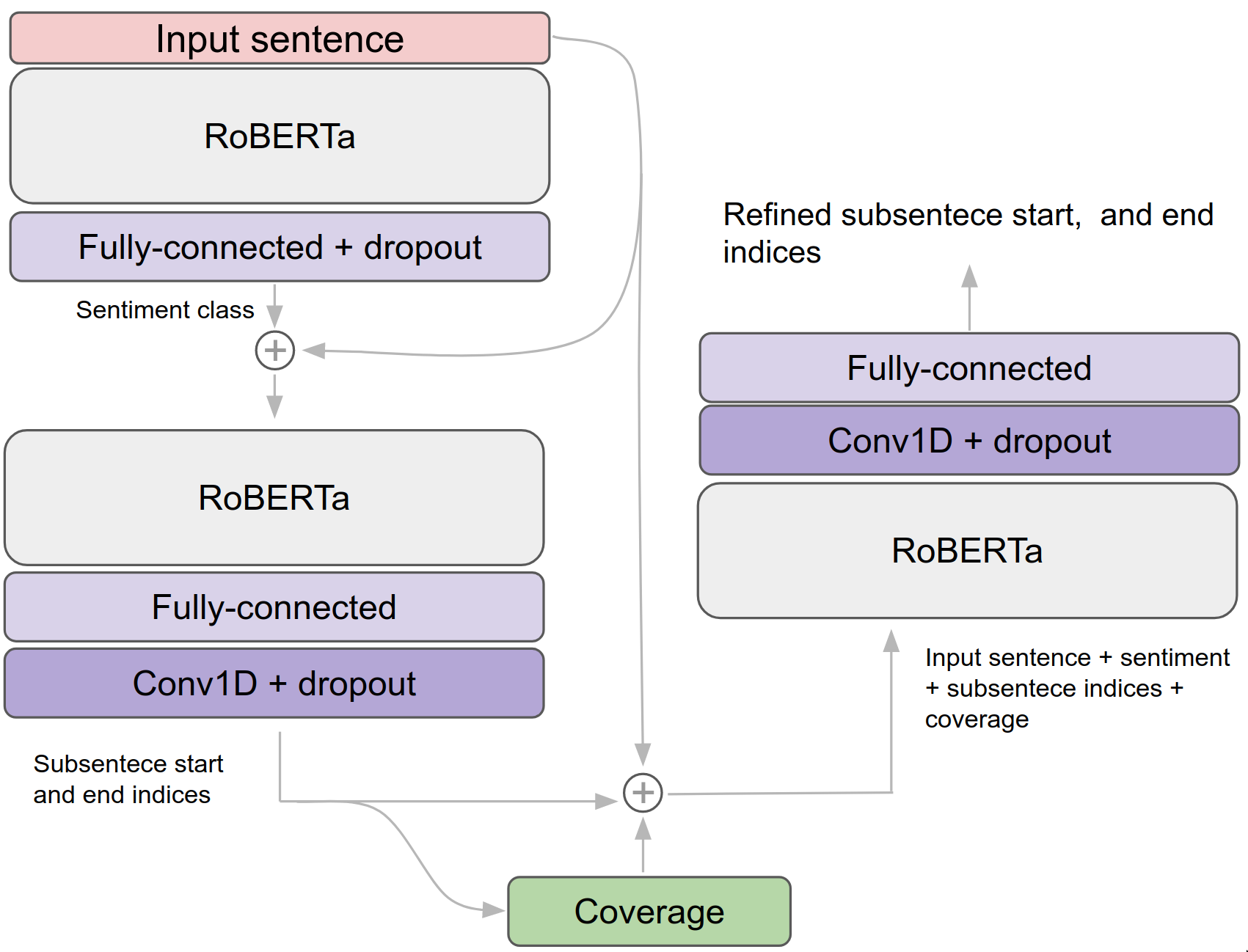}
\caption{Coverage-base subsentence extraction network architecture and pipeline. This is in-depth representation of Figure \ref{fig:pipeline} with data flow. }
\label{fig:coverageModel}
\end{figure}

\subsection{End-to-end sentiment and subsentence extraction pipeline}
Given classifier and coverage models from the previous sections, we propose an end-to-end pipeline capable of simultaneously extracting sentiment and subsentence from an input text. 
The pipeline architecture is straightforward and is formed by a series of models implying that each model's outputs are fed to the following subsequent model's inputs. We present more detail and discussion on the experimental results in section \ref{sec:pipeline}.



\section{Dataset and experiments design}
\label{sec:dataset}
In this section, we provide explanations of the dataset we used and the design of our experiments. Furthermore, input data preprocessing and tokenisation details are presented with Exploratory Data Analysis.

\begin{figure}
\centering
\includegraphics[width=\textwidth]{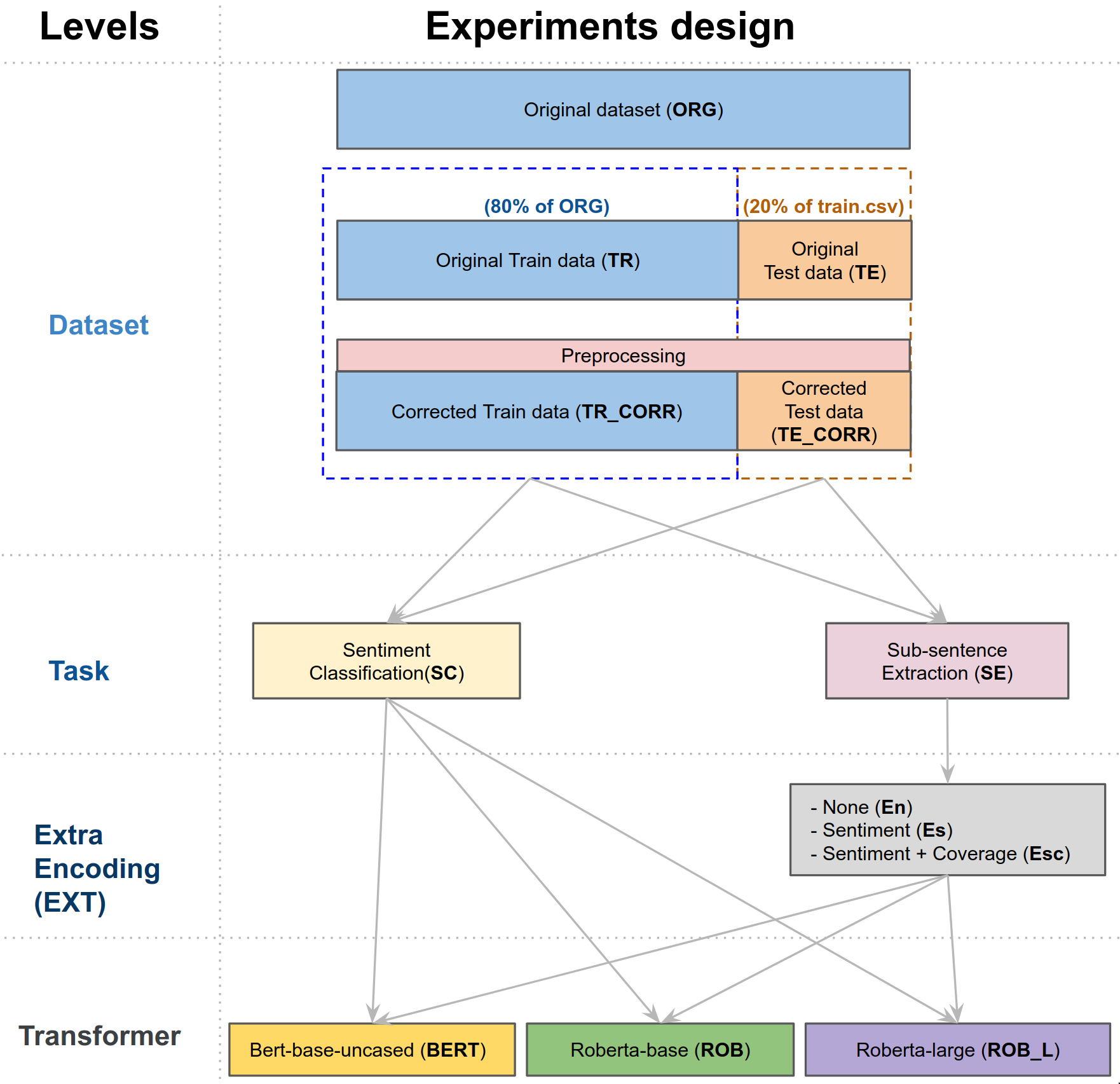}
\caption{Experiments design. This diagram shows the experiment design scheme proposed in this manuscript. There are total 24 experiments naming as the following convention: [Dataset]\_[Task]\_[EXT]\_[Transformer]. For example, 21.[TR\_CORR]\_[SE]\_[Esc]\_[ROB] indicates the experiment for Subsentence extraction task with sentiment + coverage options trained on the corrected train data with RoBERTa-base model. The prefixed number is a unique experiment ID for convenient.}
\label{fig:experiments design}
\end{figure}

As shown in Figure \ref{fig:experiments design}, we define four levels; dataset, task, extra encoding, and transformer. 
Dataset level refers to how we split and organise, train, validate and test the dataset. At high-level perspectives, we use 80\% of the original dataset to train and 20\% for the test. Note that we utilised five folds stratified Cross-Validation (CV) of the train set, meaning that the validation set takes about 16\% (i.e., one fold) and the actual training set is about 64\%. More detail regarding this, is in section \ref{sec:kfoldCrossValidation}. We define acronyms to stand for train and test dataset as \textbf{TR} and \textbf{TE}. The postfix \textbf{CORR} indicates the corrected dataset by preprocessing.

Task level has two categories; sentiment classification (\textbf{SC}) and subsentence extraction (\textbf{SE}). While the classification is directly connected to a transformer model, \textbf{SE} has three extra encoding variations; None (\textbf{En}), Sentiment (\textbf{Es}), and Sentiment with coverage (\textbf{Esc}). \textbf{En} indicates subsentence extraction without any extra encoding data, \textbf{Es} and \textbf{Esc} perform the task with sentiment and sentiment + coverage meta data respectively.

Finally, the transformer level has three variations that we introduced in the earlier section; \textbf{BERT}, \textbf{ROB} (12 encoders, 110\unit{M} parameters), and \textbf{ROB\_L} (24 encoders, 330\unit{M} parameters).

These four levels create 24 unique combinations as shown in Table \ref{tbl:exps} with the naming conventions such that [Dataset]\_[Task]\_[EXT]\_[Transformer]. For instance, 15.[TR]\_[SE]\_[Esc]\_[ROB\_L] implies the experiment for subsentence extraction task with sentiment + coverage options trained on the original training dataset with RoBERTa-large model. These notations will be consistently used across the paper.

\begin{table}
\caption{All experiments name and description. Experiment naming convention used across all sections is [Dataset]\_[Task]\_[EXT]\_[Transformer].}
\begin{tabular}{lcccc}
\toprule
\multicolumn{1}{c}{\textbf{Experiment name}}               & \textbf{Dataset}     & \textbf{Task}                                                     & \textbf{Encoding}                                              & \textbf{Transformer} \\ \hline
1.{[}TR{]}\_{[}SC{]}\_{[}BERT{]}                    & Original train data  & Classification                                                    & N/A                                                            & BERT                 \\
2.{[}TR{]}\_{[}SC{]}\_{[}ROB{]}                      & Original train data  & Classification                                                    & N/A                                                            & RoBERTa-base         \\
3.{[}TR{]}\_{[}SC{]}\_{[}ROB\_L{]}                   & Original train data  & Classification                                                    & N/A                                                            & RoBERTa-large        \\ \hline
4.{[}TR\_CORR{]}\_{[}SC{]}\_{[}BERT{]}               & Corrected train data & Classification                                                    & N/A                                                            & BERT                 \\
5.{[}TR\_CORR{]}\_{[}SC{]}\_{[}ROB{]}                & Corrected train data & Classification                                                    & N/A                                                            & RoBERTa-base         \\
6.{[}TR\_CORR{]}\_{[}SC{]}\_{[}ROB\_L{]}             & Corrected train data & Classification                                                    & N/A                                                            & RoBERTa-large        \\ \hline
7.{[}TR{]}\_{[}SE{]}\_{[}En{]}\_{[}BERT{]}           & Original train data  & \begin{tabular}[c]{@{}c@{}}Sub-sentence \vspace{-1mm} \\  Extraction\end{tabular} & None                                                           & BERT                 \\
8.{[}TR{]}\_{[}SE{]}\_{[}Es{]}\_{[}BERT{]}           & Original train data  & \begin{tabular}[c]{@{}c@{}}Sub-sentence \vspace{-1mm} \\ Extraction\end{tabular} & Sentiment                                                      & BERT                 \\
9.{[}TR{]}\_{[}SE{]}\_{[}Esc{]}\_{[}BERT{]}          & Original train data  & \begin{tabular}[c]{@{}c@{}}Sub-sentence \vspace{-1mm} \\ Extraction\end{tabular} & \begin{tabular}[c]{@{}c@{}}Sentiment + \vspace{-1mm} \\ coverage\end{tabular} & BERT                 \\ \hline
10.{[}TR{]}\_{[}SE{]}\_{[}En{]}\_{[}ROB{]}           & Original train data  & \begin{tabular}[c]{@{}c@{}}Sub-sentence \vspace{-1mm} \\ Extraction\end{tabular} & None                                                           & RoBERTa-base         \\
11.{[}TR{]}\_{[}SE{]}\_{[}Es{]}\_{[}ROB{]}           & Original train data  & \begin{tabular}[c]{@{}c@{}}Sub-sentence \vspace{-1mm} \\ Extraction\end{tabular} & Sentiment                                                      & RoBERTa-base         \\
12.{[}TR{]}\_{[}SE{]}\_{[}Esc{]}\_{[}ROB{]}          & Original train data  & \begin{tabular}[c]{@{}c@{}}Sub-sentence \vspace{-1mm} \\ Extraction\end{tabular} & \begin{tabular}[c]{@{}c@{}}Sentiment + \vspace{-1mm} \\ coverage\end{tabular} & RoBERTa-base         \\ \hline
13.{[}TR{]}\_{[}SE{]}\_{[}En{]}\_{[}ROB\_L{]}        & Original train data  & \begin{tabular}[c]{@{}c@{}}Sub-sentence \vspace{-1mm} \\ \vspace{1mm} Extraction\end{tabular} & None                                                           & RoBERTa-large        \\
14.{[}TR{]}\_{[}SE{]}\_{[}Es{]}\_{[}ROB\_L{]}        & Original train data  & \begin{tabular}[c]{@{}c@{}}Sub-sentence \vspace{-1mm} \\ Extraction\end{tabular} & Sentiment                                                      & RoBERTa-large        \\
15.{[}TR{]}\_{[}SE{]}\_{[}Esc{]}\_{[}ROB\_L{]}       & Original train data  & \begin{tabular}[c]{@{}c@{}}Sub-sentence \vspace{-1mm} \\ Extraction\end{tabular} & \begin{tabular}[c]{@{}c@{}}Sentiment + \vspace{-1mm} \\ coverage\end{tabular} & RoBERTa-large        \\ \hline
16.{[}TR\_CORR{]}\_{[}SE{]}\_{[}En{]}\_{[}BERT{]}    & Corrected train data & \begin{tabular}[c]{@{}c@{}}Sub-sentence \vspace{-1mm} \\ Extraction\end{tabular} & None                                                           & BERT                 \\
17.{[}TR\_CORR{]}\_{[}SE{]}\_{[}Es{]}\_{[}BERT{]}    & Corrected train data & \begin{tabular}[c]{@{}c@{}}Sub-sentence \vspace{-1mm} \\ Extraction\end{tabular} & Sentiment                                                      & BERT                 \\
18.{[}TR\_CORR{]}\_{[}SE{]}\_{[}Esc{]}\_{[}BERT{]}   & Corrected train data & \begin{tabular}[c]{@{}c@{}}Sub-sentence \vspace{-1mm} \\ Extraction\end{tabular} & \begin{tabular}[c]{@{}c@{}}Sentiment + \vspace{-1mm} \\ coverage\end{tabular} & BERT                 \\ \hline
19.{[}TR\_CORR{]}\_{[}SE{]}\_{[}En{]}\_{[}ROB{]}     & Corrected train data & \begin{tabular}[c]{@{}c@{}}Sub-sentence \vspace{-1mm} \\ Extraction\end{tabular} & None                                                           & RoBERTa-base         \\
20.{[}TR\_CORR{]}\_{[}SE{]}\_{[}Es{]}\_{[}ROB{]}     & Corrected train data & \begin{tabular}[c]{@{}c@{}}Sub-sentence \vspace{-1mm} \\ Extraction\end{tabular} & Sentiment                                                      & RoBERTa-base         \\
21.{[}TR\_CORR{]}\_{[}SE{]}\_{[}Esc{]}\_{[}ROB{]}    & Corrected train data & \begin{tabular}[c]{@{}c@{}}Sub-sentence \vspace{-1mm} \\ Extraction\end{tabular} & \begin{tabular}[c]{@{}c@{}}Sentiment + \vspace{-1mm} \\ coverage\end{tabular} & RoBERTa-base         \\ \hline
22.{[}TR\_CORR{]}\_{[}SE{]}\_{[}En{]}\_{[}ROB\_L{]}  & Corrected train data & \begin{tabular}[c]{@{}c@{}}Sub-sentence \vspace{-1mm} \\ Extraction\end{tabular} & None                                                           & RoBERTa-large        \\
23.{[}TR\_CORR{]}\_{[}SE{]}\_{[}Es{]}\_{[}ROB\_L{]}  & Corrected train data & \begin{tabular}[c]{@{}c@{}}Sub-sentence \vspace{-1mm} \\ Extraction\end{tabular} & Sentiment                                                      & RoBERTa-large        \\
24.{[}TR\_CORR{]}\_{[}SE{]}\_{[}Esc{]}\_{[}ROB\_L{]} & Corrected train data & \begin{tabular}[c]{@{}c@{}}Sub-sentence \vspace{-1mm} \\ Extraction\end{tabular} & \begin{tabular}[c]{@{}c@{}}Sentiment +\vspace{-1mm} \\ coverage\end{tabular} & RoBERTa-large \\ \hline
\bottomrule
\end{tabular}
\label{tbl:exps}
\end{table}

\subsection{Tokenisation and input data preprocessing}

This section describes the input data preprocessing idea and tokenisation that we performed before model training.

The preprocessing objective is to uniformly format and clean input data so that the subsequent tokeniser and model will be able to interpret efficiently. Therefore, preprocessing is one of the important procedures of NLP for training well-generalised and high-performance language models.
There are many preprocessing techniques such as lower-casing, removing unnecessary characters (extra white-spaces, special characters, HTML tags), or lemmatisation (stemming of words, e.g., cars or car's will be converted to car). To address all possible cases is a non-trivial task due to high-complexity and large diversity in natural language, thus we simply yet effectively applied lowercase to all texts and removed URL and HTML tags from the original input training data. In addition, the original dataset contains meaningless full stops (e.g., "..." or ".."), which are trivial in sentimental perspectives but significantly affect sentiment classification results and subsentence extraction performance. Therefore we replaced them with a single stop. We will discuss more of this topic in the experimental section.

Followed by special character cleaning, we also performed tokenisation that converts words into either characters or subwords, as it is considered one of the most important steps in NLP. In essence, the idea is to split the word into meaningful pieces and map between an input word and the corresponding digits (so-called encoding). This is shown in Table \ref{tbl:token}. For example, from the Figure, we have an input sentence, ``Hello this is a really good wine'' with the subsentence that is a manual label justifying why this sentence is labelled as a positive sentiment. Then, \textbf{tokens} and \textbf{input\_ids} show tokenised words and the corresponding encoding. Note that there is a specific input format of RoBERTa, which is originally designed for a question and answer task such as starting with <s> tag followed by question tokens and adding stop tags with answer tokens at the end. For the \textbf{attention\_mask} we set all as true because we want to the model to focus on all tokens except padding. \textbf{start\_token} and \textbf{end\_token} indicate the start and end index of the selected text. RoBERTa makes use of GPT-2 tokeniser, using byte-level BPE \cite{sennrich2015neural} and we utilise the popular pre-trained RoBERTa tokeniser from huggingface\footnote{\url{https://huggingface.co/transformers/model_doc/roberta.html}}.


\tabcolsep=1pt
\begin{table}[]
\caption{An exemplified input data and its tokenisation}
\begin{tabular}{rcccccccccccccc}
\toprule
\multicolumn{1}{l}{Input sentence =} & \multicolumn{5}{l}{\begin{tabular}[c]{@{}l@{}}"Hello this is a \\ really good wine"\end{tabular}} & \multicolumn{2}{l}{Subsentence =} & \multicolumn{2}{l}{"really good"}   & \multicolumn{1}{l}{}        & \multicolumn{1}{l}{Sentiment =} & \multicolumn{1}{l}{positive} & \multicolumn{1}{l}{}         & \multicolumn{1}{l}{}         \\ \hline
tokens =                             & \textless{}s\textgreater{}          & Hello          & this          & is            & a          & really           & good           & wine  & \textless{}/s\textgreater{} & \textless{}/s\textgreater{} & positive                        & \textless{}/s\textgreater{}  & \textless{}pad\textgreater{} & \textless{}pad\textgreater{} \\
input\_ids =                         & 0                                   & 812            & 991           & 2192          & 12         & 3854             & 202            & 19292 & 2                           & 2                           & 1029                            & 2                            & 1                            & 1                            \\
attention\_mask =                    & 1                                   & 1              & 1             & 1             & 1          & 1                & 1              & 1     & 1                           & 1                           & 1                               & 1                            & 0                            & 0                            \\
start\_token =                       & 0                                   & 0              & 0             & 0             & 0          & 1                & 0              & 0     & 0                           & 0                           & 0                               & 0                            & 0                            & 0                            \\
end\_token =                         & 0                                   & 0              & 0             & 0             & 0          & 0                & 1              & 0     & 0                           & 0                           & 0                               & 0                            & 0                            & 0                    \\ \hline
\bottomrule
\end{tabular}
\label{tbl:token}
\end{table}




\subsection{Tweet sentiment dataset}

We used a public dataset from the Kaggle tweet sentiment extraction competition\footnote{\url{https://www.kaggle.com/c/tweet-sentiment-extraction/data}}. Among many other publicly available datasets such as ISEAR\footnote{\url{https://www.kaggle.com/shrivastava/isears-dataset}}, or SemEval-2017\footnote{\url{https://alt.qcri.org/semeval2017/task4/}}, this dataset is unique because it contains not only sentiment of sentences (e.g., positive, negative, or neutral) but more importantly `selected text' which are words or phrases drawn from original tweets. This `selected text' is why the corresponding sentence holds its sentiment, as shown in Table \ref{tbl:tr-data}. The total number of samples in this dataset is about 27\unit{k} and detail is provided in Table \ref{tbl:dataset}.




\begin{table}[H]
\caption{Tweet sentiment dataset}
\centering
\begin{tabular}{c|ccccc}
\toprule
\multicolumn{1}{l|}{}  & \textbf{Total} & \textbf{Unique} & \textbf{Positive} & \textbf{Negative} & \textbf{Neutral} \\
\midrule
\textbf{Tweet}         & 27480          & 27480           & 8582(31.23\%)     & 7781(28.32\%)     & 11117(40.45\%)   \\
\textbf{Selected text} & 27480          & 22463           & N/A               & N/A               & N/A \\ \hline
\bottomrule
\end{tabular}
\label{tbl:dataset}
\end{table}

Table \ref{tbl:tr-data} shows the top 5 row-wise samples from the dataset (i.e., each row indicates one instance). Each row has five columns; `textID' is an unique ID for each piece of text, `text' is the text of the tweet, `selected text' is the text that supports the tweet's sentiment, and `sentiment' is the general sentiment of the tweet. We further discuss the dataset in terms of Exploratory Data Analysis (EDA) and data preparation for model training in the following sections.

\subsection{Exploratory Data Analysis (EDA)}
Nowadays, there is well-established knowledge and resource in machine learning or data-driven approaches such as highly-engineered frameworks; Tensorflow, PyTorch, or advanced-tools; SciPy\footnote{\url{https://www.scipy.org/}} or scikit-learn\footnote{\url{https://scikit-learn.org/}}. Their high-fidelity and powerful feature extraction capability may lead to outstanding performance without any data analysis (just feeding all training data and waiting). However, a high-level understanding of the dataset often plays a crucial role in improving model performance and spotting missing and erroneous data.  

In handling sequential information (i.e., in our sentence processing case), correlating words' contextual meaning and their distribution in training and test sets must be identified prior to model training. Hence, in this section, we provide our EDA strategies and share some underlying insights of the dataset used.

We exploited a public dataset from a tweet-sentiment-extraction competition\footnote{\url{https://www.kaggle.com/c/tweet-sentiment-extraction}}. The organiser provided 27k training samples composed of raw tweet sentence, selected text, and sentiment, as shown in Table \ref{tbl:tr-data}.


\begin{table}[H]
\caption{5 samples from training dataset with meta information.}
\centering
\begin{tabular}{@{}lllll@{}}
\toprule
           & \textbf{textId} & \textbf{text}                                    & \textbf{selected\_text}             & \textbf{sentiment} \\ \midrule
\textbf{0} & cb774db0d1      & I'd have responded, if I were going              & I'd have responded, if I were going & neutral            \\
\textbf{1} & 549e992a42      & Sooo SAD I will miss you here in San Diego!!!    & Sooo SAD                            & negative           \\
\textbf{2} & 088c60f138      & my boss is bullying me...                        & bullying me                         & negative           \\
\textbf{3} & 9642c003ef      & what interview! leave me alone                   & leave me alone                      & negative           \\
\textbf{4} & 358bd9e861      & Sons of ****, why couldn't they put them on t... & Sons of ****,                       & negative           \\ 
\hline
\bottomrule
\end{tabular}
\label{tbl:tr-data}
\end{table}

The first column is a unique textID, but this does not provide any useful information. The second and third columns are original tweets and manually labelled texts that determine the type of sentiment in the last column. For example, the second sample, ``Sooo SAD I will miss you here in San Diego!!!'' is labelled as negative sentiment because of ``Sooo SAD'' texts. Note that these selected texts (or ground truth) are manually labelled which can be often subjective, and the organiser randomly sampled training and test sets with some extra noise (e.g., random white spaces in labelling). Although these may result in performance degrading, the dataset is unique and has a sufficiently large volume of samples, so we decided to use this data for model training.


\begin{table}[H]
\caption{Examples of incorrect and corrected selected text (right most column). selected\_text and sentiment columns are manually labelled meta information which are corrupted due to special characters.}
\centering
\begin{tabular}{@{}llllll@{}}
\toprule
\textbf{text}                                       & \textbf{selected\_text} & \textbf{sentiment} & \textbf{corrected\_selected\_text} \\ \midrule
is back home now gonna miss every one               & onna                    & negative           & miss                               \\
He's awesome... Have you worked with him before...  & s awesome               & positive           & awesome.                           \\
hey mia! totally adore your music. when ...         & y adore                 & positive           & adore                              \\
Nice to see you tweeting! It's Sunday 10th...       & e nice                  & positive           & nice                               \\
\#lichfield \#tweetup sounds like fun Hope to...    & p sounds like fun       & positive           & sounds like fun                    \\
nite nite bday girl have fun at concert             & e fun                   & positive           & fun                                \\
HaHa I know, I cant handle the fame! and thank you! & d thank you!            & positive           & thank you!                         \\ 
\hline
\bottomrule
\end{tabular}
\label{tbl:corrected}
\end{table}

\begin{figure}
\centering
\includegraphics[width=0.8\textwidth]{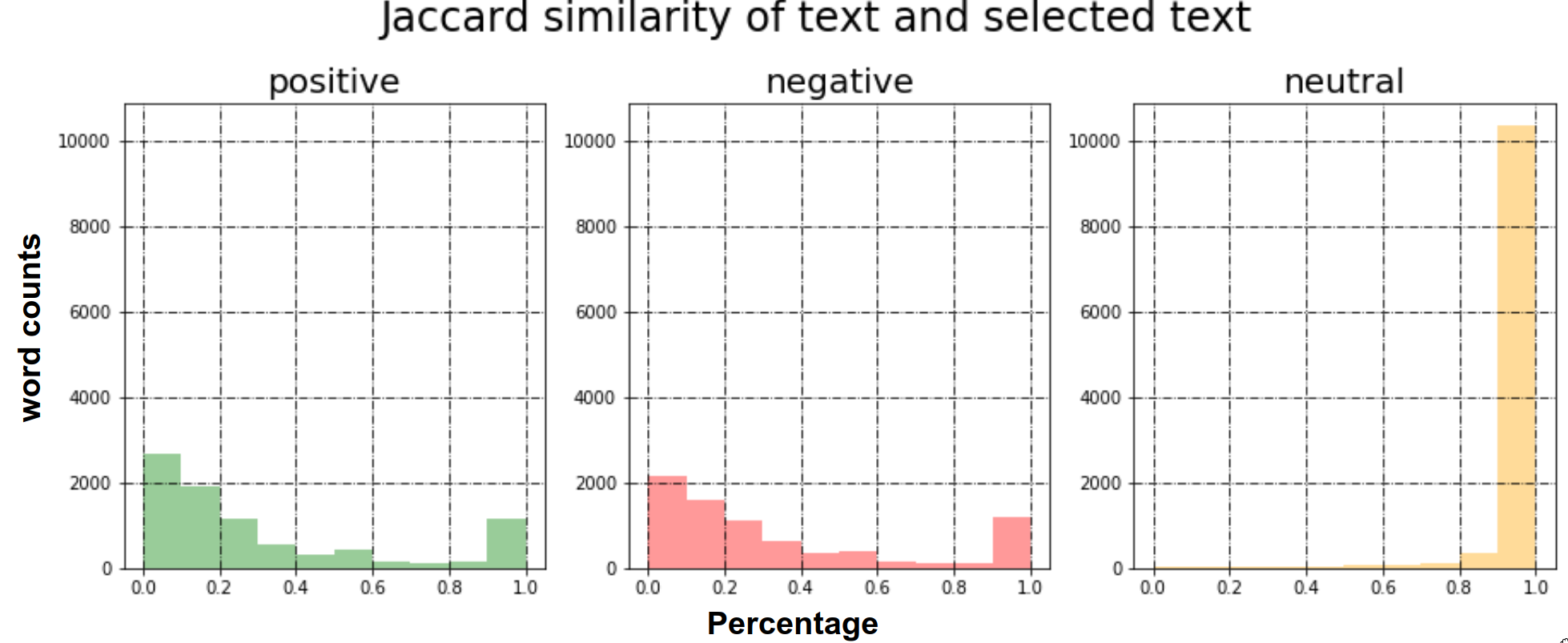}
\caption{These plots display Jaccard similarity between input sentence and selected subsentence for each sentiment in the dataset. x-axis is the similarity percentage and y-axis is the number of sentences holding the corresponding similarity. Noticeably, sentences with neutral sentiment have the identical subsentence whereas sentences with positive and negative show similar distributions.}
\label{fig:jaccard-text-sel}
\end{figure}

It is important to note that there are noises or artifacts in the training dataset. For example, the top row from Table \ref{tbl:corrected}, has ``onna'' as the selected\_text with negative sentiment. This sounds irrelevant, and further investigation found that this happens due to $N$ exceeding and $M$ leading spaces surrounding the selected\_text. The original text of the top row is ``is\textvisiblespace back\textvisiblespace home\textvisiblespace now\textvisiblespace\textvisiblespace\textvisiblespace\textvisiblespace\textvisiblespace\textvisiblespace      gonna\textvisiblespace miss\textvisiblespace every\textvisiblespace one'' and we can now see the six invisible leading spaces prior to ``gonna'' and our corrected selected text should be ``miss' after removing these six leading spaces. We explicitly calculated $N$ and $M$ and found that there were 1112 cases where selected\_text mismatching happened in positive and negative samples (i.e., 14.7\%). Note that we did not correct neutral samples since their text and selected\_text are almost identical, as shown in Figure \ref{fig:jaccard-text-sel}. 



This corrected dataset will be exploited as the baseline dataset across this manuscript, and we will revisit this later in section \ref{sec:coverage}. The similarity metric used in this paper (i.e., Jaccard score, and Area-Under-the Curve, AUC) will be presented in the following experiment results section.



\begin{figure}
\centering
\includegraphics[width=0.8\textwidth]{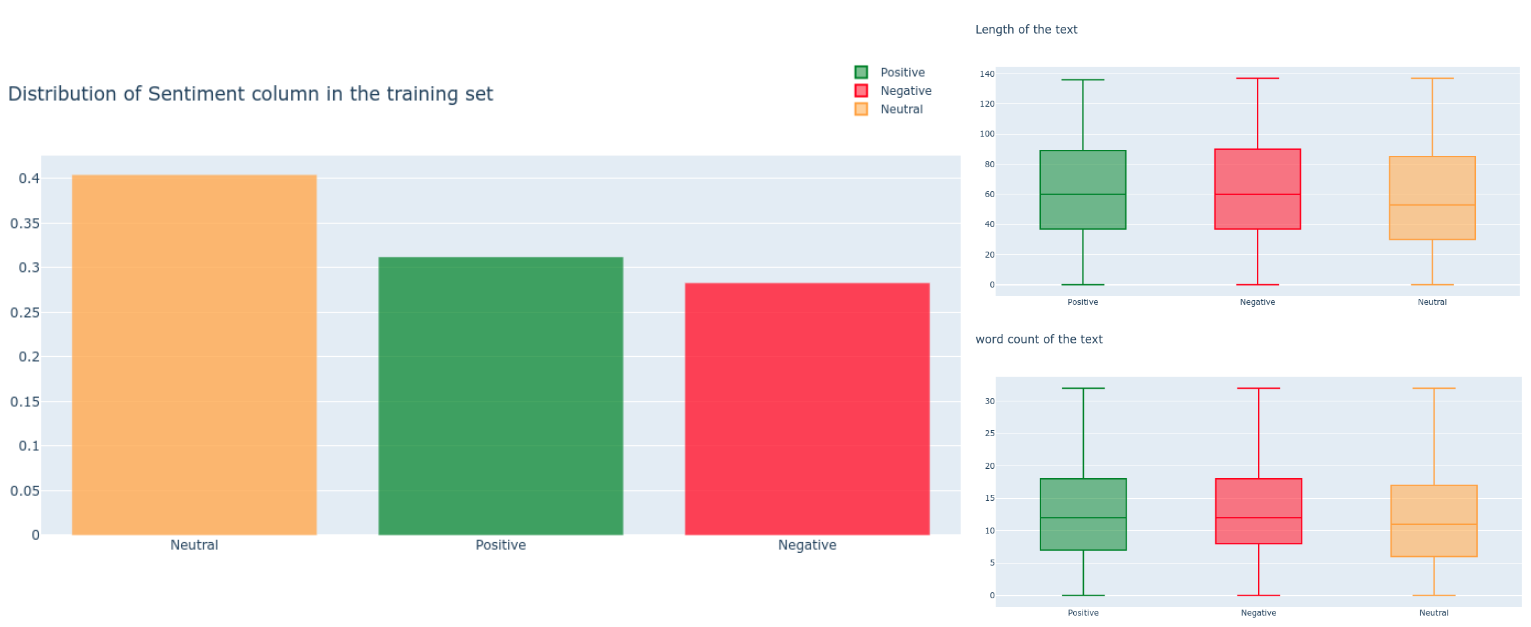}
\caption{Left illustrates sentiment distributions and right is length of text and word count for each sentiment.}
\label{fig:eda1}
\end{figure}

It is observed that there is a trivial class unbalancing between positive and negative samples (see Figure \ref{fig:eda1}). Within a deep-learning context, we think this margin is negligible and that these can be considered as well-balanced samples. Additionally, the length of sentence and words are not useful features in distinguishing sentiment. 

\begin{figure}
\centering
\includegraphics[width=0.6\textwidth]{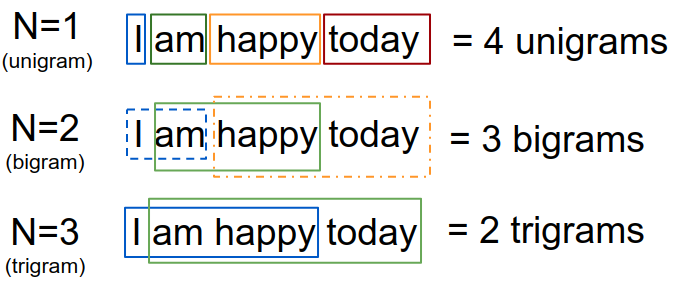}
\caption{Examples of $N$-grams. Each box groups words depending on $N$.}
\label{fig:n-grams}
\end{figure}

Other statistics that we can extract from the corpus are N-grams that are a contiguous sequence of N words/tokens/items from a given sample of text, as illustrated in Figure \ref{fig:n-grams}. 

An input sentence, `I am happy today' produces four unigrams (`I', `am', `happy', `today') and three bigrams (`I am', `am happy', `happy today'). The unigrams themselves may be insufficient to capture context, but bigrams or trigrams contain more meaningful and contextually rich information. Based on this observation, we see what the most appearing N-grams in our dataset are.

\begin{figure}
\centering
\includegraphics[width=0.8\textwidth]{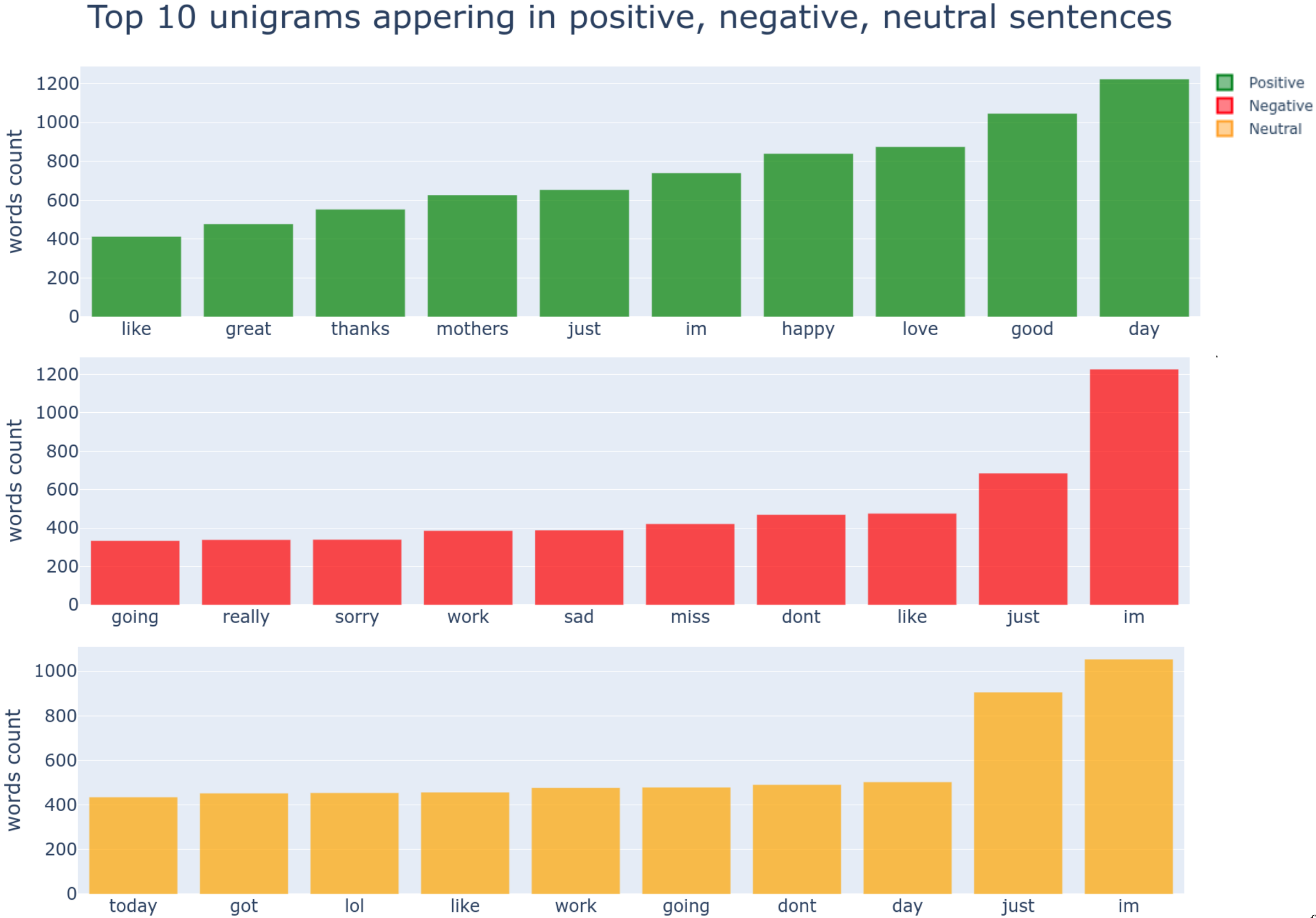}
\caption{Statistic of top 10 unigrams in positive, negative, neutral sentiment. `im` and `just` appear in all sentiments with high counts implying that leveraging unigrams meta information may cause confusion for models and lead to inferior performances.}
\label{fig:unigrams}
\end{figure}

\begin{figure}
\centering
\includegraphics[width=0.8\textwidth]{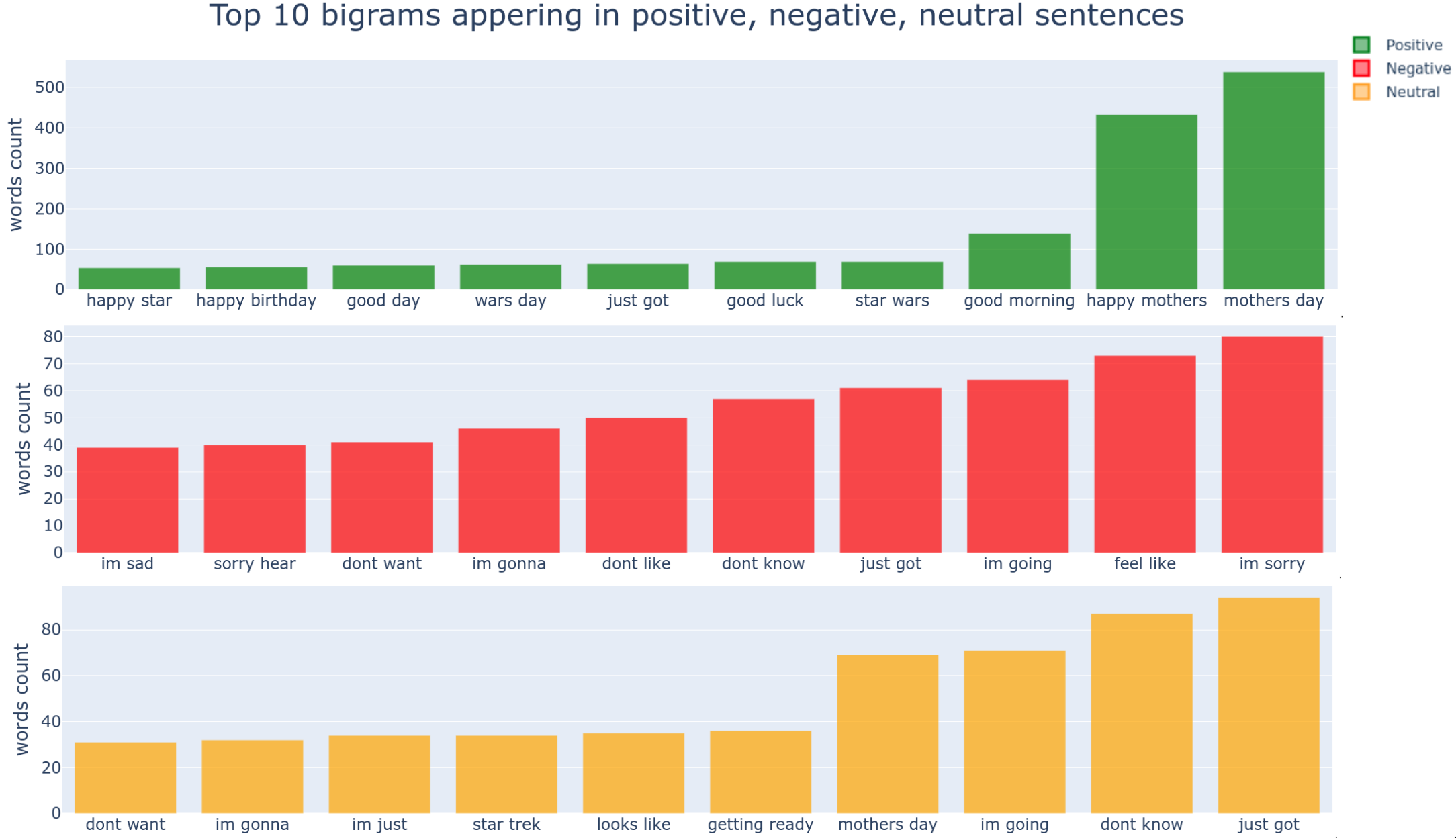}
\caption{Top 10 bigrams appearing in positive, negative, neutral sentiment. By looking at top tier bigrams, they demonstrate better discrimination between sentiment than unigram.}
\label{fig:bigrams}
\end{figure}

Figure \ref{fig:unigrams} and \ref{fig:bigrams} show the frequency of uni-, bigrams in our training dataset. From the unigrams, one can find that positive words such as `good', `great', `like', `happy', `love' ranked near the top and negative words; `sad', `sorry', `dont' are top-ranked. These results are expected results because the words in the same sentiment are closely located in the word embedding space. However, it is rather difficult to gauge solely based on unigrams (e.g., `im' appears in all sentiments).

The bigrams or trigrams can encapsulate a more distinguishable meaning, as shown in Figure \ref{fig:bigrams}. It is noticeable that `mothers day' is top-ranked, and we guess this tweet dataset was sampled around 9th/May.

From the EDA study, we conclude that the number of words or the length of sentences has a marginal impact in extracting sentiment, but the most important features are a contiguous sequence of words that align with our approach presented in section \ref{sec:coverage}.

\subsection{Dataset preparation}
\label{sec:datasetPreparation}
\begin{figure}
\centering
\includegraphics[width=\textwidth]{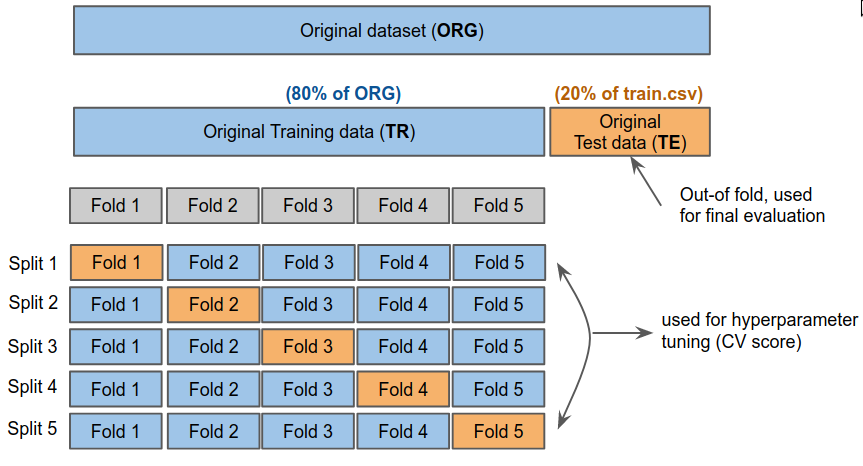}
\caption{Five folds cross-validation dataset split\protect\footnotemark. We exploit this validation strategy across all experiments. Blue boxes are training set, and orange indicates validation/test set.}
\label{fig:data_split}
\end{figure}

In this section, we present a dataset split strategy for model training and validation. As shown in Figure~\ref{fig:experiments design}, we split the original dataset as 80\% train and the rest, 20\% for testing. Although this split ratio is empirically chosen, it is commonly used in NLP and image related deep learning tasks. Figure~\ref{fig:data_split} illustrates train/test split and five folds CV split. We present further technical detail regarding CV in the following section \ref{sec:kfoldCrossValidation}. The blue boxes are the training set, and the orange indicates the testing set. All 24 experiments presented in this paper follow this dataset split rule.

\footnotetext{reproduced of \url{https://scikit-learn.org/stable/modules/cross_validation.html}}

\section{Model Training}
\label{sec:modelTraining}

The object of this section is training all 24 experiments (see Table \ref{tbl:exps}). Given the the prepared dataset, this section presents the details of model training strategies such as Stratified K-fold cross-validation and training configuration. 

\subsection{Model details}
\label{sec:modelDetails}
We train three models; BERT, RoBERTa-base, and RoBERTa-large, for two different tasks; classification and coverage-based subsentence extraction. As mentioned earlier in section \ref{sec:bert-roberta}, we used the original transformer architectures of BERT and RoBERTa and attached varying head layers that are placed on the top of each transformer, depending on the task. For the classification task, a fully-connected layer (input dim: 768, output dim: 3) with a dropout (0.1) added, followed by the transformer model as shown in Table~\ref{tbl:arch}. The output of this network indicates how likely the input sentence is classified as one of the predefined sentiment classes (e.g., positive, negative, and neutral). This prediction is useful and boosts the subsentence extraction performance significantly (more detail regarding this is presented in the following section). This is mainly because the classification network is able to guide the subsentence extraction network toward the direction where it extracts subsentences from more correlated distributions. For example, suppose an input sentence is classified as positive. In that case, the subsentence extraction network is likely looking for positive words (e.g., good, or fantastic) rather than negative words (bad, or poor). In turn, this also leads to narrowing down the overall search space that improves convergence time. 

Coverage-based subsentence extraction network has slightly more complex architecture as shown in Figure~\ref{fig:coverageModel}. The inputs of this are input sentence, classified sentiment, and coverage that indicates the percentage of subsentence from the entire input sentence. The outputs are the indices of the starting and end words within the input sentence. We added three 1D Convolutional layers, followed by two fully-connected layers. Model hyperparameters are empirically selected, and details can be found in Table~\ref{tbl:arch}.

\begin{table}[]
\caption{Emotion classification and subsentence extraction network architecture}
\begin{threeparttable}
\begin{tabular}{ccc|ccc}
\toprule

\multirow{2}{*}{Task} & \multirow{2}{*}{Layer}   & Input          & \multirow{2}{*}{Task}                                                                               & \multirow{2}{*}{Layer}   & Input                                                                                    \\
                      &                          & Output         &                                                                                                     &                          & Output                                                                                   \\ \hline \hline
Classification        & \multirow{2}{*}{RoBERTa} & Input sentence & \multirow{2}{*}{\begin{tabular}[c]{@{}c@{}}Covereage-based\\ subsentence\\ extraction\end{tabular}} & \multirow{2}{*}{RoBERTa} & \begin{tabular}[c]{@{}c@{}}Input + Attention Mask \\ + Sentiment + Coverage\end{tabular} \\
                      &                          & 768            &                                                                                                     &                          & 768                                                                                      \\ \cline{2-3} \cline{5-6} 
                      & Dropout                  & 0.1            &                                                                                                     & Dropout                  & 0.3                                                                                      \\ \cline{2-3} \cline{5-6} 
                      & \multirow{2}{*}{FC}\tnote{a}      & 768            & \multirow{2}{*}{}                                                                                   & \multirow{2}{*}{Conv1D}  & 768                                                                                      \\
                      &                          & 3              &                                                                                                     &                          & 256                                                                                      \\ \cline{2-3} \cline{5-6} 
                      &                          &                &                                                                                                     & \multirow{2}{*}{Conv1D}  & 256                                                                                      \\
                      & \multirow{2}{*}{}        &                &                                                                                                     &                          & 128                                                                                      \\ \cline{5-6} 
                      &                          &                &                                                                                                     & \multirow{2}{*}{Conv1D}  & 128                                                                                      \\
                      &                          &                &                                                                                                     &                          & 64                                                                                       \\ \cline{5-6} 
                      &                          &                &                                                                                                     & \multirow{2}{*}{FC}      & 64                                                                                       \\
                      &                          &                &                                                                                                     &                          & 32                                                                                       \\ \cline{5-6} 
                      &                          &                &                                                                                                     & \multirow{2}{*}{FC}      & 32                                                                                       \\
                      &                          &                &                                                                                                     &                          & 2 \\ \hline
\bottomrule
\end{tabular}
\begin{tablenotes}[para]
    \item[a] Fully-connected
\end{tablenotes}
\end{threeparttable}
\label{tbl:arch}
\end{table}

It is important to note that this coverage-based model makes use of the outputs of other networks such as sentiment and indices predictions. This is explained in more detail in the end-to-end entire pipeline section (section~\ref{sec:pipeline}). 

\subsection{K-fold Cross Validation}
\label{sec:kfoldCrossValidation}
It is common to use $K$-fold cross-validation (CV) for model validation where K implies a model is trained on $K - 1$ folds and tested with one remaining fold. Although this process is expensive (because it enforces $K$ times model training phases), it is considered one of the most data efficient and acceptable validation methods by machine learning communities. We thus apply five stratified CV\footnote{\url{https://scikit-learn.org/stable/modules/generated/sklearn.model_selection.StratifiedKFold.html}} to all 24 models, to ensure properly equalised class distributions between training and test set of each fold.

\subsection{Training Configuration}
\label{sec:trainConfiguration}
All model training was conducted on an RTX3090 GPU that has 24\unit{GB}. BERT, and RoBERTa-base took approximately one hour, and RoBERTa-large finished five folds CV in two hours. In total, it took about 31 hours for the full model training. Training multiple models with such variants opens a tricky task because there are many things to consider, such as model parameters, logging, and validation. We have developed a unified and automated NLP framework, BuilT\footnote{\url{https://github.com/UoA-CARES/BuilT}}, that takes care of all necessary house-keeping and experiments above. Using BuilT, one can reproduce the same results we presented in this paper and develop other similar projects.

Adam optimiser (lr = 0.00003) is used with categorical Cross-Entropy Loss (CEL) \cite{goodfellow2016deep} of smoothed label:

\begin{equation}
y^{LS} = y(1 - \alpha) + \frac{\alpha}{C}
\end{equation}

where $\alpha$ is smoothness (e.g., when $\alpha$ is 0, then it is identical to CEL) and $C$ is the number of label classes (our case $C$ == 3). With the label smoothing, our loss is defined as:

\begin{align}
CE(x_i, y_i^{LS}) =& - \sum_{i = 1}^{C}y_i^{LS} \log h_{\theta}(x_i)
\end{align}
where $h_{\theta}(x_i)$ indicates the softmax output of our model parameterised by $\theta$ given $i$th input $x_i$ and is defined as
\begin{align}
h_{\theta}(x_i) =& \frac{\exp(x_i)}{\sum_{j=1}^{C}\exp(x_i)}
\end{align}
Finally, CEL can be compuated as 
\begin{align}
CEL(x_i, y_i^{LS}) =& -\sum_{j = 1}^{N}(\sum_{i = 1}^{C}y_i^{LS} \log \frac{\exp(x_i)}{\sum_{i=1}^{C}\exp(x_i)})
\end{align}
where $N$ is the total number of training samples.

\begin{figure}
\centering
\includegraphics[width=0.8\textwidth]{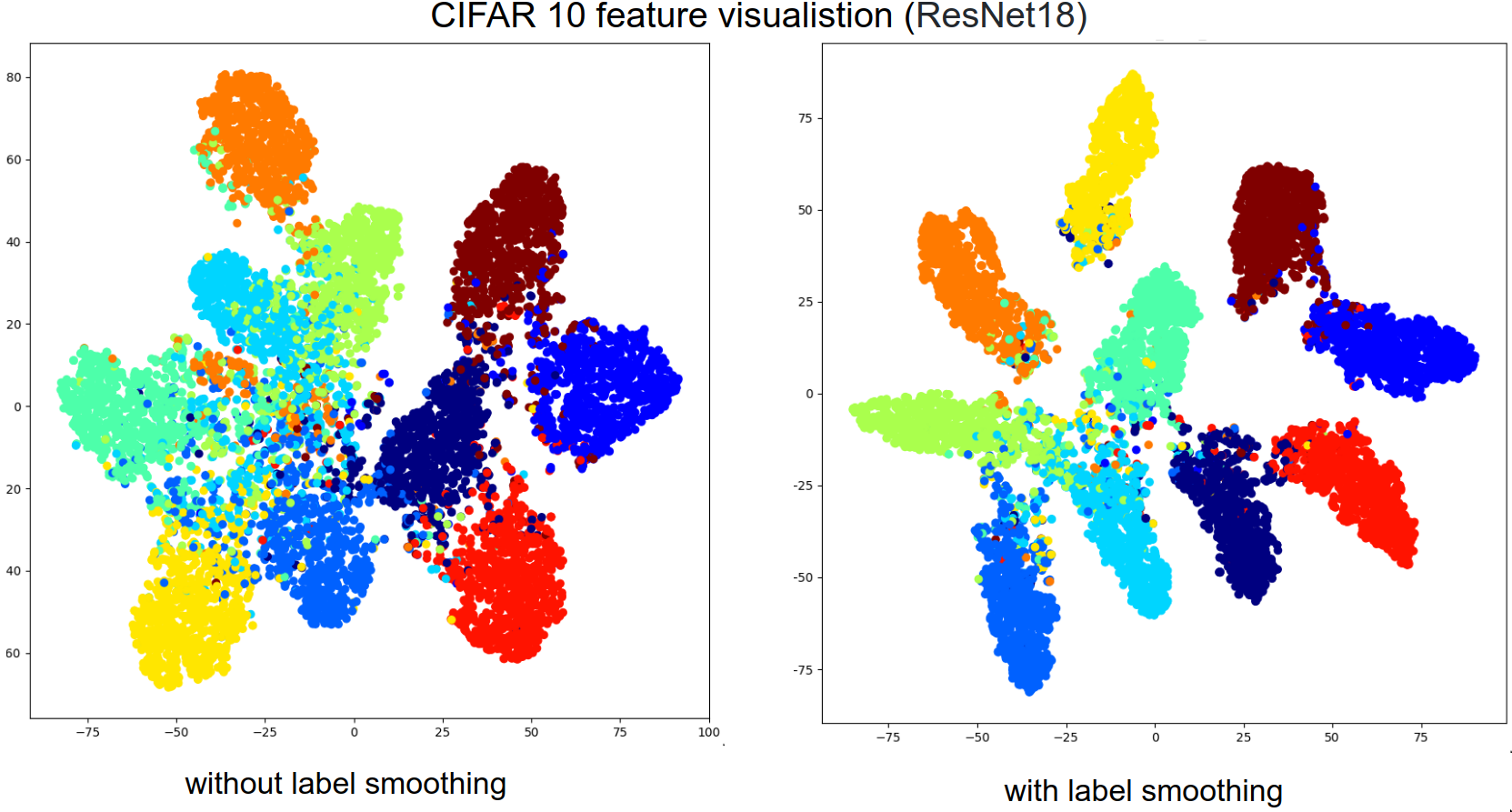}
\caption{Feature visualisation using TSNE \cite{Van_der_Maaten2008-yg} without (left) and with label smoothing for a classification task. Clearer inter-class boundaries can be observed with label smoothing technique.}
\label{fig:label-smoothing}
\end{figure}

The use of this label smoothing \cite{goodfellow2016deep} as exemplified in Figure~\ref{fig:label-smoothing} improved performance significantly in our case. This is mainly due to the fact that there is possible noise in the manual label (i.e., training set) so that this may force the model to learn inaccurate label strictly. Label smoothing eases the label as a function of $\alpha$, making it less critical to this incorrect label. Note that this method does not always lead to superior results, and it depends on the quality of the dataset. Multistep learning rate scheduler ($\gamma = 0.1$, stepping at 3, 4, 5) was used across all experiments.

\section{Experimental Results}
\label{sec:results}
In this section, evaluation metrics used for evaluating the model performances (Jaccard score, AUC, and $F_1$ scores) are defined, followed by classification and coverage-based subsentence extraction models. Finally, the entire pipeline model evaluation is conducted for real-world applications and usage.

\subsection{Evaluation metrics}
\label{sec:evalMetrics}
We used three metrics for different tasks because an individual task requires a task-specific metric. For the classification task, AUC and $F_1$ scores were used. These metrics can capture both precision and recall performance, and are commonly accepted for the task. For the subsentence extraction task, Jaccard score is utilised for measuring the similarity between subsentences.

\subsubsection{Area-Under-the-Curve}
AUC is the metric we used for gauging the performance of our classifier. As the name suggests, it measures the total area under the receiver operating characteristic curve (i.e., precision-recall curve). Precision ($P$) and recall ($R$) can be expressed as

\begin{equation}
P = \frac{T_{\mbox{\tiny{P}}}}{T_{\mbox{\tiny{P}}} + F_{\mbox{\tiny{P}}}}, \;\;\;R = \frac{T_{\mbox{\tiny{P}}}}{T_{\mbox{\tiny{P}}} + F_{\mbox{\tiny{N}}}}
\end{equation}
where $T_{\mbox{\tiny{P}}}$ and $F_{\mbox{\tiny{P}}}$ are true/false positive which stand for correct/incorrect classification and $F_{\mbox{\tiny{N}}}$ is false negative, that indicates mis-clssification. From $P$ and $R$, we can construct a performance curve by varying all possible thresholds as shown in Figure \ref{fig:classification}. Since this curve encapsulates all possible thresholds, it is independent of threshold tuning and closer to 1 indicates better performance. Due to these powerful and intuitive concepts, this AUC metric is one of the most widely accepted metrics in classification problems.

\subsubsection{Harmonic $F_1$ score}
Similarly, we also evaluate the proposed system using $F_1$ score which can be expressed as

\begin{equation}
F_1 = \frac{T_{\mbox{\tiny{P}}}}{T_{\mbox{\tiny{P}}} + \frac{1}{2}({F_{\mbox{\tiny{P}}} + F_{\mbox{\tiny{N}}})}}
\end{equation}.

This harmonic $F_1$ score is one of the most widely accepted metrics in a classification task because it reflects precision and recall performance.

\subsubsection{Jaccard score for subsentence extraction}
Jaccard score is a measure of similarity between two sentences. The higher the score, the more similar the two strings. In this paper, we tried to find the number of common tokens and divide it by the total number of unique tokens. This may be expressed as
\begin{equation}
J_{\mbox{\tiny{score}}}(A,B) = \frac{|A \cap B|}{|A \cup B|} = \frac{|A \cap B |}{|A| + |B| - |A\cap B|}.
\end{equation}
As an example, if we want to calculate a $J_{\mbox{\tiny{score}}}$ between sentences A, "Hello this is a really good wine" and B, "Hello, this is a really good wine." (note that there is a trailing comma and period in sentence B), the score is 0.555. This metric is significantly lower than we anticipated based on the meanings of two input sentences (which are identical). Since we compare token level and split words based on space, 'Hello' and 'Hello,', 'wine' and 'wine.' are treated as different words.

\subsection{Model ensemble}
As mentioned in section~\ref{sec:kfoldCrossValidation}, we used five folds CV for all 24 experiments and this implies that we have five different models for each experiment. One of most popular methods to fuse these models is ensemble averaging, $\Tilde{y}(\boldsymbol{x})$ and may be expressed as
\begin{equation}
\Tilde{y}(\boldsymbol{x}) = \sum_{i=1}^{N}\alpha_{i}y_i(\boldsymbol{x}) 
\end{equation}
$N$ is the number of models to fuse (i.e., five in this case), and $\boldsymbol{x}$ is the input vector, $y_i$ is the prediction of $i$th model, and $\alpha_i$ is the weight of $i$th model. In our case, we used an equal average ensemble which implies that $\alpha_i = \frac{1}{N}$. The equally-weighted ensemble strategy is rather simple yet powerful in dealing with model overfitting and in the presence of a noisy label (e.g., model voting). There is a wide-range of ensemble options reported within the literature \cite{sagi2018ensemble} on how to select the optimal weights for a particular task. Weaknesses in the use of an ensemble is that this requires $\mathcal{O}(N)$ time complexity for sequential processing or $\mathcal{O}(N)$ space for holding models in a machine's memory.

Table \ref{tbl:ensemble} shows equally weighted model ensemble results. Most of experiments report performance boost. However, as depicted by model 23 (bottom row), it is not always guaranteed to obtain superior performance with ensembled model. In such case, other ensemble options can be applied.

\begin{table}[H]
\centering
\caption{Equally weighted ensemble result summary}
\begin{threeparttable}
\begin{tabular}{ccc}
\toprule
\multicolumn{1}{c}{}                                          & \multicolumn{2}{c}{Classification}                                      \\
\multicolumn{1}{c}{Trained Model from Experiment}                                     & Average F1                          & Ensemble F1                          \\ \hline
\multicolumn{1}{c}{13.{[}TR\_CORR{]}\_{[}SC{]}\_{[}BERT{]}$\uparrow$\tnote{a}}   & 0.7890                           & 0.7957                               \\
\multicolumn{1}{c}{14.{[}TR\_CORR{]}\_{[}SC{]}\_{[}ROB{]}$\uparrow$}    & 0.7921                           & 0.8026                               \\
\multicolumn{1}{c}{15.{[}TR\_CORR{]}\_{[}SC{]}\_{[}ROB\_L{]}$\uparrow$} & 0.7973                           & \textbf{0.8084}                      \\
                                                              & \multicolumn{1}{l}{}             & \multicolumn{1}{l}{}                 \\
                                                              & \multicolumn{2}{l}{Subsentence extraction}                    \\
\multicolumn{1}{c}{Trained Model from Experiment}                                     & \multicolumn{1}{l}{Average Jaccard} & \multicolumn{1}{l}{Ensemble Jaccard} \\ \hline
17.{[}TR\_CORR{]}\_{[}SE{]}\_{[}En{]}\_{[}ROB{]}$\uparrow$              & 0.6457                           & 0.6474                               \\
20.{[}TR\_CORR{]}\_{[}SE{]}\_{[}Es{]}\_{[}ROB{]}$\uparrow$              & 0.7249                           & 0.7257                               \\
23.{[}TR\_CORR{]}\_{[}SE{]}\_{[}Esc{]}\_{[}ROB{]}$\uparrow$             & \textbf{0.8944}                  & 0.8900       \\ \hline
\bottomrule
\end{tabular}
\begin{tablenotes}[para]
    \item[a] the closer to 1 indicates the better performance
\end{tablenotes}
\end{threeparttable}
\label{tbl:ensemble}
\end{table}

\subsection{Classification results}
\label{sec:classificaiton-results}
Precise sentiment prediction is an essential front-end component by providing the scope of the subsentence search space. The goal of this section is to choose the most suitable sentiment classification network via intensive performance evaluation.

\begin{figure}
\centering
\includegraphics[width=\textwidth]{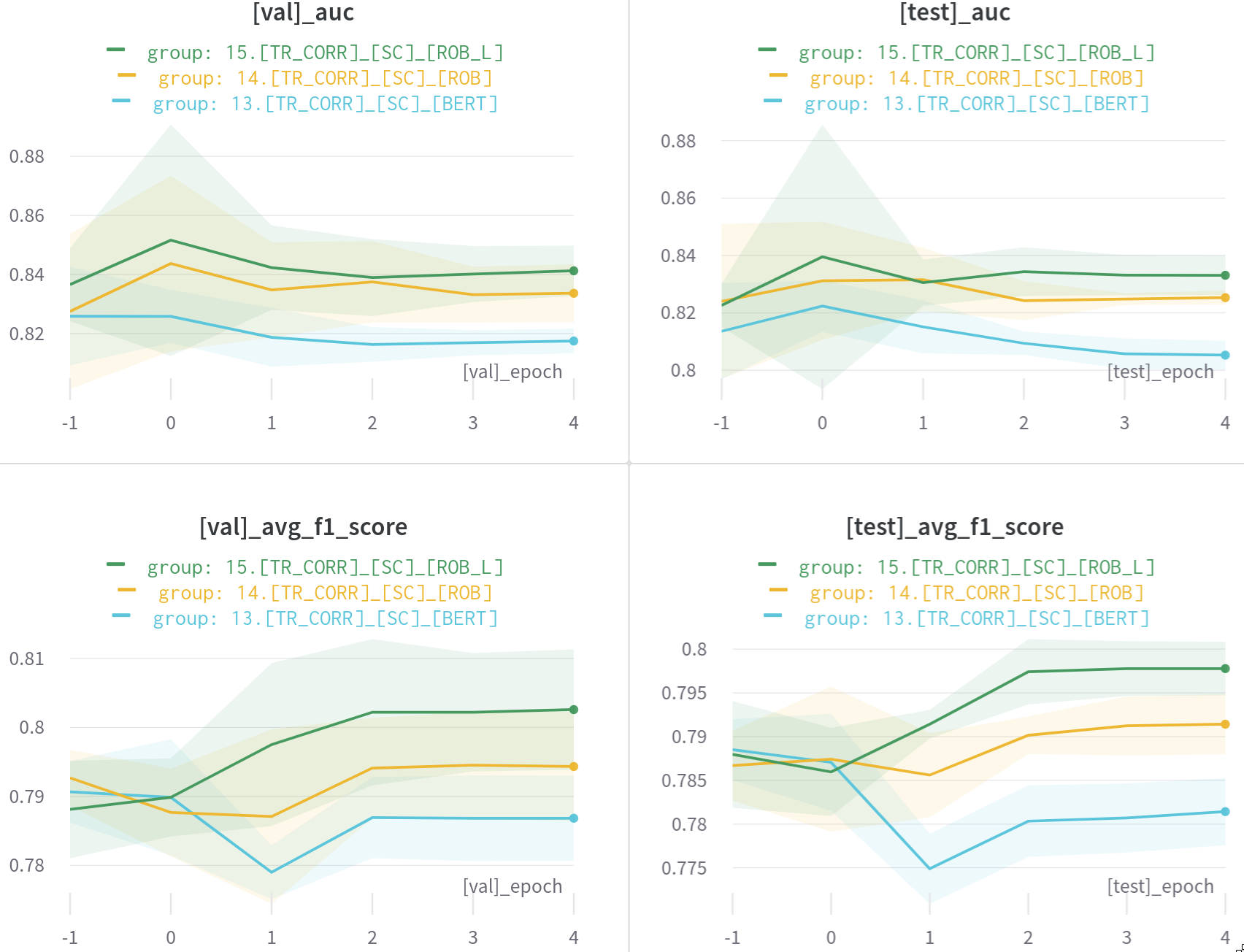}
\caption{Index extraction results (AUC and F1 score) of three different models (i.e., Bert, RoBERTa base, and RoBERTa large. Shaded area is the standard deviation of 5 models from cross-validation and solid lines denote average metrics displayed top of each plot.}
\label{fig:classification}
\end{figure}

We performed six experiments with variations in datasets and models, and decided to use the RoBERTa-base model. For visual inspection purposes, we only present three experiments in this section. The summary is shown in Table~\ref{tbl:classificationResults}, and experimental results are displayed in Figure~\ref{fig:classification}. The horizontal axis indicates epoch, and the vertical axis is the corresponding performance metric. The shaded area is the stand deviation of five models with the mean value indicated by the solid line. We followed model naming conventions as defined Table~\ref{tbl:exps} in section~\ref{sec:dataset}. As an instance, [TR\_CORR]\_[SC]\_ROB] can be interpreted by the RoBERTa-base model trained on the corrected dataset for the sentiment classification task.

Overall, the RoBERTa-large model performed best for all validation and test experiments. However, this network consists of twice as many encoders as RoBERTa-base, which means theoretically about two times slower inferencing time. Our experiments prove that BERT and RoBERTa-base took about 4\unit{ms} per sentence (100\unit{s} for processing 5.4\unit{K} test sentences) whereas 8\unit{ms} for RoBERTa-large (220\unit{s}). However, this is marginal to our real-time sentiment extraction task considering processing time and subsequent tasks such as SSML and multilabel extraction. Furthermore, the performance degradation with RoBERTa-base is 0.05$\sim$0.1, which lies within our acceptable range. Therefore, we decided to use RoBERTa-base network for the following the end-to-end pipeline \ref{sec:pipeline}.


\begin{table}[H]
\caption{Sentiment classification results summary}
\centering
\begin{threeparttable}
\begin{tabular}{ccccccc}
\toprule
         & \multicolumn{2}{c}{13.{[}TR\_CORR{]}\_{[}SC{]}\_{[}BERT{]}} & \multicolumn{2}{c}{14.{[}TR\_CORR{]}\_{[}SC{]}\_{[}ROB{]}\tnote{a}} & \multicolumn{2}{c}{15.{[}TR\_CORR{]}\_{[}SC{]}\_{[}ROB\_L{]}} \\
         & Val                          & Test                         & Val                          & Test                        & Val                                   & Test                        \\ \hline
Average AUC$\uparrow$ & 0.8175                       & 0.8053                       & 0.8336                       & 0.8253                      & \textbf{0.8412}                       & 0.8331                      \\
Average $F_1$$\uparrow$  & 0.7868                       & 0.7978                       & 0.7943                       & 0.7914                      & \textbf{0.8026}                       & 0.7814                    \\ \hline
\bottomrule
\end{tabular}
\begin{tablenotes}[para]
  \item[a] This is the model used for the subsequent coverage model
  \end{tablenotes}
  \end{threeparttable}
  \label{tbl:classificationResults}
\end{table}

\subsection{Coverage-based subsentence extraction}
\begin{figure}
\centering
\includegraphics[width=\textwidth]{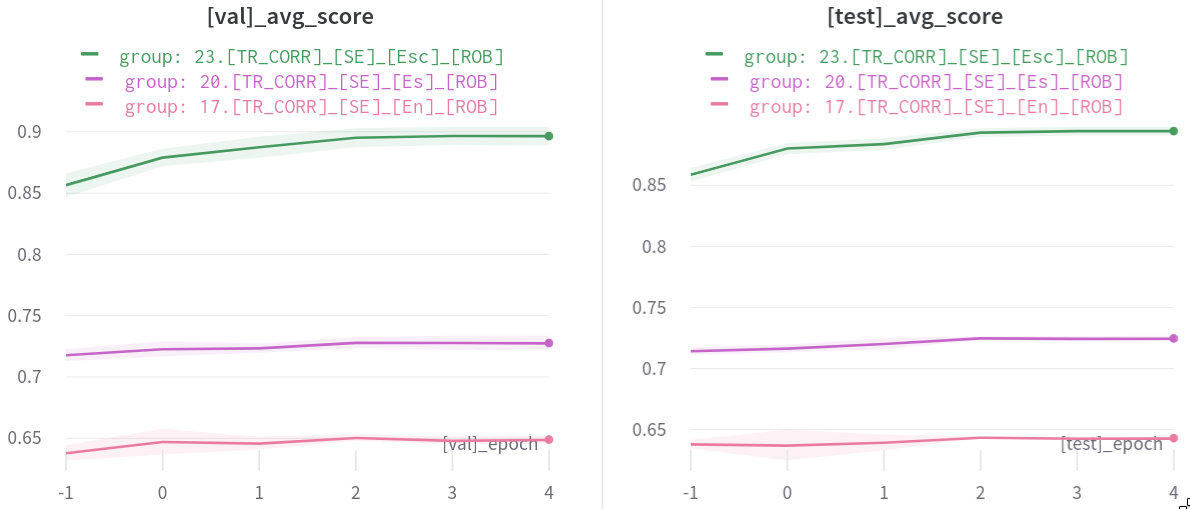}
\caption{Jaccrad score performance evaluation of the proposed coverage model (green). Shaded area is the standard deviation of 5 models. Model 17 (pink, with no additional meta data) performs least but Model 20 (magenta) and 23 (green) demonstrate impressive performance improvements with meta information.}
\label{fig:coverage}
\end{figure}

Given the sentiment classification and subsentence extraction information, our coverage model aims to improve the performance by selectively refining span predictions. As shown in Figure~\ref{fig:coverage}, the coverage model outperforms other models with a large margin.

17.{[}TR\_CORR{]}\_{[}SE{]}\_{[}En{]}\_{[}ROB{]} model indicates performing subsentence extraction without sentiment information with RoBERTa-base backbone and model 20 is with that metadata. The presence of sentiment data boosts the overall Jaccard score by about 0.8. It is obvious that the extra encoding of data helped the model better localise correct indices (i.e., start and end indices of subsentence). Additional coverage information (model 23) increased the performance to around 0.89, which impressively demonstrates our proposed approach. Overall a summary is provided in Table~\ref{tbl:coverageSummaryResults}. It is important to note that the model 23 assumes that sentiment information is also given, which is not always true in real-world scenarios where only an input sentence is provided. We are then required to predict not only coverage, but sentiment as well in the pipeline. Therefore, we present the pipeline experiment results in the following section.

\begin{table}[H]
\caption{Coverage-based subsentence extraction results summary}
\centering
\begin{tabular}{ccc}
\toprule
\multicolumn{1}{c}{Model}                         & \multicolumn{1}{l}{Val. Jaccard} & \multicolumn{1}{l}{Test Jaccard} \\ \hline
17.{[}TR\_CORR{]}\_{[}SE{]}\_{[}En{]}\_{[}ROB{]}  & 0.649                            & 0.643                            \\
20.{[}TR\_CORR{]}\_{[}SE{]}\_{[}Es{]}\_{[}ROB{]}  & 0.7277                           & 0.7247                           \\
23.{[}TR\_CORR{]}\_{[}SE{]}\_{[}Esc{]}\_{[}ROB{]} & \textbf{0.8965}                  & \textbf{0.8944}  \\ \hline
\bottomrule
\end{tabular}
\label{tbl:coverageSummaryResults}
\end{table}

\subsection{End-to-end pipeline evaluation results}
\label{sec:pipeline}
\begin{figure}
\centering
\includegraphics[width=\textwidth]{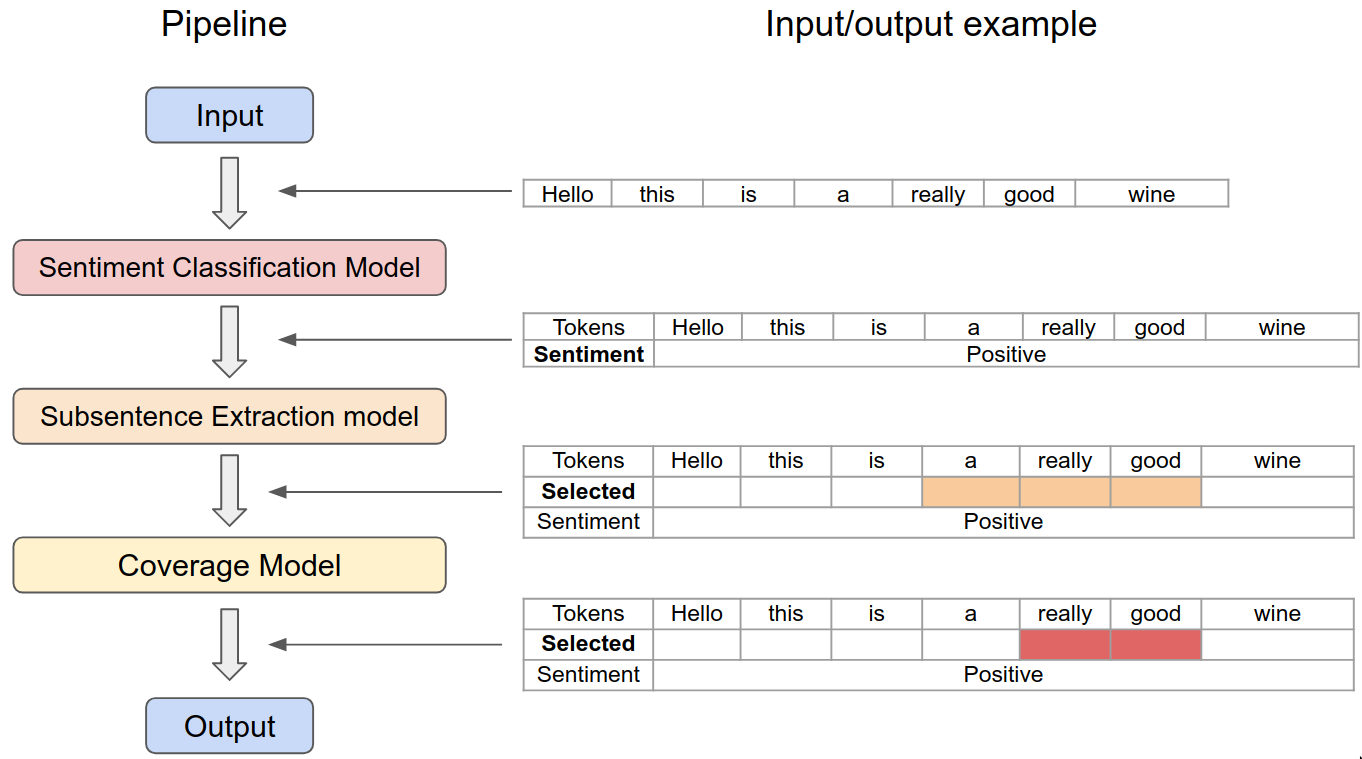}
\caption{Sentiment and subsentence prediction pipeline. This diagram is manually illustrated with an example just to highlight refining procedure with the proposed coverage model.}
\label{fig:pipeline-results}
\end{figure}

In order to use the proposed approach in real-world scenarios such as HRI, one must perform sentiment and subsentence predictions sequentially. Based on previous experiments, we developed a pipeline predicting sentiment using 14.[TR\_CORR]\_[SC]\_[ROB] and feed the sentiment to 20.[TR\_CORR]\_[SE]\_[Es]\_[ROB] for the initial subsentence extraction and finally, 23.[TR\_CORR]\_[SE]\_[Esc]\_[ROB] for refined subsentence prediction. Figure~\ref{fig:pipeline-results} illustrates the pipeline with an example input sentence. In this example, it is observed that our coverage model can refine the subsentence range so that it leads to superior results.

We tested the proposed pipeline with four cases conditioned by a sentiment and coverage model, as shown in table~\ref{tbl:pipelineSummaryResults}. From the table, the left column indicates the source of sentiment meta information and the top row is the presence of the coverage model. The model with coverage and without it for inferencing 5.4\unit{k} sentences achieved Jaccard scores of \textbf{$0.7265$}, and $0.7256$ respectively. The proposed coverage model slightly improves the overall performance of both cases. Note that the result shown in the middle row (i.e., Model 20) is inferior to the bottom that assumes the sentiment label is given from the manual label. This is caused by error propagation in the sentiment model throughout the entire pipeline. 

\begin{table}[H]
\caption{Pipeline subsentence extraction results summary}
\centering
\begin{tabular}{ccc}
\toprule
\multicolumn{1}{c}{Source of sentiment}                         & \;\;\;With coverage & \;\;\;W/O coverage \\ \hline
Classifier prediction  & \textbf{0.6401}                            & 0.6392                            \\
Manual label  & \textbf{0.7265}                           & 0.7256                           \\ \hline
\bottomrule
\end{tabular}
\label{tbl:pipelineSummaryResults}
\end{table}




\subsection{Class Activation Mapping}
\begin{figure}
\centering
\includegraphics[width=\textwidth]{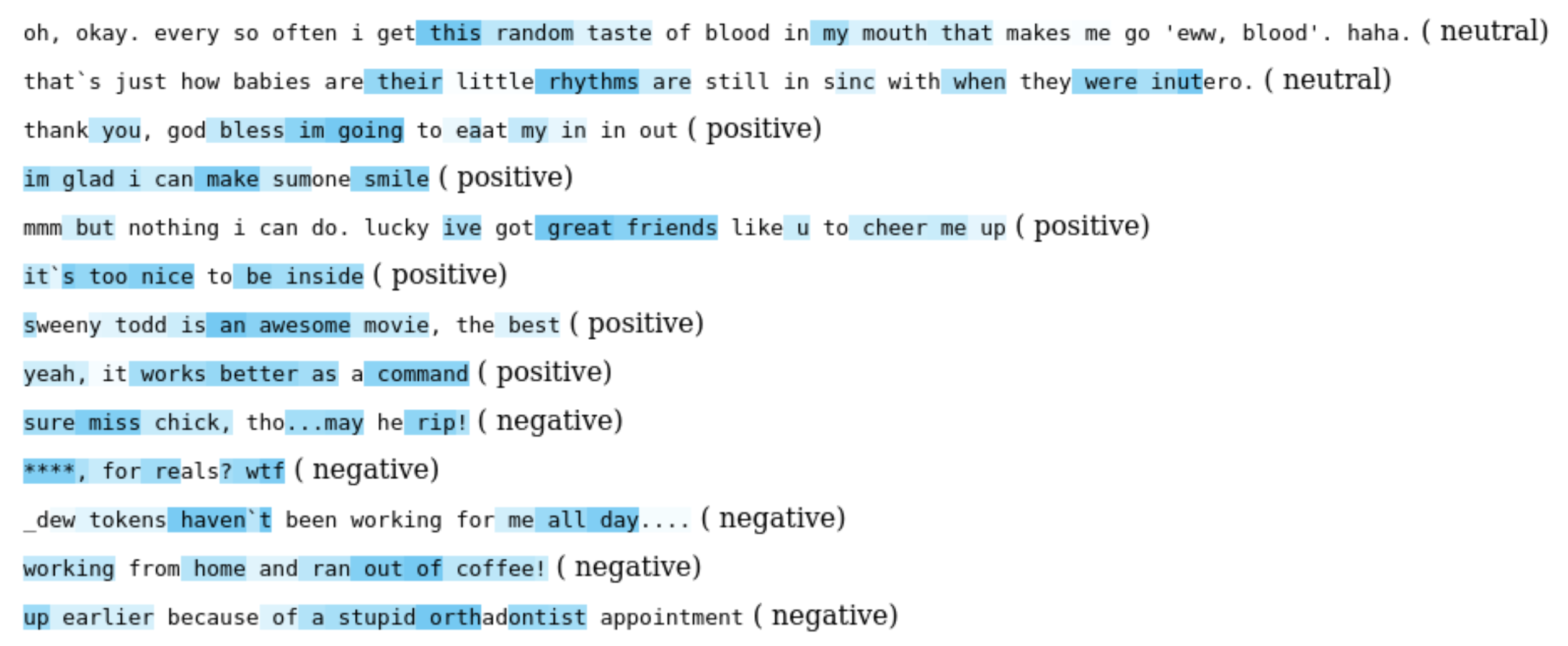}
\caption{Examples of multilabel attention. Each row is an input sentence with highlighted predicted activation. The darker blue represents the higher probability. It is important to note that our framework is capable of predicting multiple labels from one sentence which can result in better machine-human interaction experience.}
\label{fig:multilabel-extraction}
\end{figure}

In performing the sentiment classification task, networks often only output classes with the corresponding probability. However, they also learned about what caused that output in their internal layers. For example, in the image classification task, an activation map can be extracted from one of the internal layers that highlight a region of interest in an image. This technique is useful for debugging purposes and, more importantly, valuable feature extraction. Analogous to this, in a NLP context, we can use a Class Activation Mapping (CAM) technique to extract the activation of a word that indicates how much the word contributed to the classification task. Figure~\ref{fig:multilabel-extraction} illustrates activation of words identifying contribution to sentiment classification. For example, the words `smile`, `awesome` and `great` show stronger responses (the darker blue is the higher probability) in the positive sentence. Another useful aspect is that multi-activation can be leveraged as information-rich features or input for speech synthesis markup language (SSML) for more vivid and enhanced natural human-machine interaction.

\section{Discussion and limitations}
\label{sec:discuss}
This section reports the issues and limitations that we have encountered while developing the sentiment and subsentence extraction framework and when conducting the experiments.

We encountered a common machine learning problem; high-variance or overfitting. Even though the dataset is well-balanced between positive and negative sentiment, neutral takes a large portion of the dataset that we bypassed for subsentence extraction (i.e., an input sentence with neutral sentiment is identical to its subsentence). This operation was intentionally performed due to the dataset reflecting this pattern. In turn, this affects the coverage model to avoid shrinking its coverage range. This is why our coverage model has a lack of ability to shrink, rather than expanding. Another reason may be that the average length of subsentences in the dataset is relatively long with respect to input sentence (e.g., $\geq0.5$) so that the model could only learn to expand a coverage span. This depends on how the dataset labelled by human labellers.
To address this, we are investigating adaptive coverage span methods by making use of different means such as a weighted cross-entropy loss function that treats each class with varying weights that are estimated from class distributions. In addition, introducing outlier rejection that filters out samples that have large margins with respect to their class distributions.

\section{Conclusions}
\label{sec:conclusions}
In this paper, we propose a real-time and high accurate sentence to sentiment framework. Based upon the recent success in NLP using transformer models, we designed a coverage-based subsentence extraction model that outperforms other work with a large margin. This model learns how to further refine a span of subsentence in a recursive manner, and in turn, increases performance. We intensively evaluated the performance of 24 models by utilising acceptable metrics and methods such as AUC, and $F_1$ scores for classification and Jaccard for subsentence extraction and stratified five folds CV. The use of meta-information significantly helps models predict accurate subsentences (e.g., 0.643 to 0.724 Jaccard score), and coverage encoding impressively improves performance by a large margin (0.8944). We also demonstrate that an ensemble of five models boosts the performance and present experiments with the entire pipeline framework that takes only a sentence as input to be used for many real-world useful applications.

\vspace{6pt} 



\authorcontributions{``conceptualization, J.L. and I.S.; methodology, J.L. and I.S; software, J.L. and I.S; validation, J.L. and I.S; resources, J.L. and I.S; writing--original draft preparation, J.L. and I.S; writing--review and editing, J.L., I.S, H.A., N.G, and S.L; visualization, J.L. and I.S; supervision, H.A. and B.M.; funding acquisition, H.A}

\funding{This work was supported by the Technology Innovation Program (10077553, Development of Social Robot Intelligence for Social Human-Robot Interaction of Service Robots) funded By the Ministry of Trade, Industry \& Energy (MI, Korea). The authors would also like to acknowledge the University of Auckland Robotics Department and the CARES team for their ongoing support during this study.}

\acknowledgments{We would like to thank Dr. Ahamadi Reza for providing useful feedback and proof-reading.}

\conflictsofinterest{The authors declare no conflicts of interest.} 

\abbreviations{The following abbreviations are used in this manuscript\\

\noindent 
\begin{tabular}{@{}ll}
AI & Artificial Intelligence \\
AUC & Area-Under-the Curve \\
BERT & Bidirectional Encoder Representations from Transformers\\
BPE & Byte-Pair Encoding \\
CAM & Class Activation Mapping \\
CEL & Cross-Entropy Loss \\
CV  & Cross-Validation \\
EDA & Exploratory Data Analysis52 \\
GLUE & General Language Understanding Evaluation \\
GPT-3 & Generative Pre-trained63Transformer 3 \\
HRI & Human-Robot Interaction \\
NLP & Natural Language Processing \\
NSP & Next Sentence Prediction \\
RoBERTa & A Robustly Optimized BERT Pretraining Approach \\
S2E & Sentence To Emotion \\
SOTA & State-Of-The-Art \\
SSML & Speech Synthesis Markup Language \\
SQuAD & Stanford Question Answering Dataset \\
\end{tabular}}




\reftitle{References}


\externalbibliography{yes}
\bibliography{main.bbl}



\end{document}